%% file: main.tex
\documentclass[manuscript,screen]{acmart}
\usepackage[vlines]{tabularht}
\usepackage{adjustbox}
\usepackage{longtable}
\usepackage{cleveref}
\usepackage{booktabs}
\usepackage{multirow}
\usepackage{siunitx}
\usepackage{pifont}
\usepackage{graphicx}
\usepackage{hyperref,url}
\hypersetup{
    colorlinks=true,
    linkcolor=blue,
    }
  
\usepackage{tablefootnote}
\usepackage{enumitem,booktabs}
\usepackage[referable]{threeparttablex}
\renewlist{tablenotes}{enumerate}{1}
\begin{document}

\title{Advancements in Molecular Property Prediction: A Survey of Single and Multimodal Approaches}

\author{Tanya Liyaqat}
\affiliation{%
  \institution{Jamia Millia Islamia}
  \city{New Delhi}
  \country{India}
}

\author{Tanvir Ahmad}
\affiliation{%
  \institution{Jamia Millia Islamia}
  \city{New Delhi}
  \country{India}}

\author{Chandni Saxena}
\affiliation{%
  \institution{The Chinese University of Hong Kong}
  \country{SAR China}}


\renewcommand{\shortauthors}{}
\begin{abstract}
Molecular Property Prediction (MPP) plays a pivotal role across diverse domains, spanning drug discovery, material science, and environmental chemistry. Fueled by the exponential growth of chemical data and the evolution of artificial intelligence, recent years have witnessed remarkable strides in MPP. However, the multifaceted nature of molecular data, such as molecular structures, SMILES notation, and molecular images, continues to pose a fundamental challenge in its effective representation. To address this, representation learning techniques are instrumental as they acquire informative and interpretable representations of molecular data. This article explores recent AI-based approaches in MPP, focusing on both single and multiple modality representation techniques. It provides an overview of various molecule representations and encoding schemes, categorizes MPP methods by their use of modalities, and outlines datasets and tools available for feature generation. The article also analyzes the performance of recent methods and suggests future research directions to advance the field of MPP.
\end{abstract}





\maketitle
\input{introduction.tex}
\input{availability_DME.tex}%
\input{encoding_schemes.tex}%
\input{modality_based_MPP.tex}
\input{nn_and_ls.tex}
\input{resource_availability.tex}
\input{comparative_analysis.tex}
\input{discussion.tex}%
\input{Conclusion.tex}



\input{main.bbl}

\input{appendix.tex}

\end{document}

%% file: introduction.tex
\section{Introduction}\label{sec:introduction}
Predicting molecular properties is a critical endeavor in drug discovery and development that poses a substantial challenge to researchers. Artificial intelligence (AI) has revolutionized various field by providing innovative solutions to the complex problems. In recent years, AI has significantly advanced molecular property prediction (MPP) by providing researchers with powerful tools to expedite the drug discovery process~\cite{shen2019molecular}. Traditionally, MPP strategies relied on expert features of molecular data to predict molecular properties.  While effective, This approach requires extensive domain knowledge to identify appropriate features for designing predictive models. Deep learning (DL) have shifted this paradigm by automatically learning intricate patterns and representations. 
This shift has democratized MPP, making the field more accessible by removing the requirement for manual feature engineering~\cite{merkwirth2005automatic}. However, molecular structures, characterized by atoms and bonds, present significant challenges for direct processing. Similarly, the sequential representation of molecules, such as SMILES strings, encounters issues related to uniqueness and length. To address these challenges, DL architectures are designed to represent molecule structures or SMILES into fixed-size vectors. 
Notable approaches for representation learning are Graph Neural Networks (GNNs), Recurrent Neural Networks (RNNs), Transformers, Convolutional Neural Networks (CNNs), and others~\cite{rong2020self,zeng2022accurate}. These methods can extracts meaningful features from molecular structures and encapsulate the intricate relationships between a molecule chemical composition and its bioactivity~\cite{yang2019analyzing}. Deep learning has also revolutionized Quantitative Structure-Activity Relationship (QSAR) modeling.
For example, Stokes et al.~\cite{stokes2020deep} used DL to virtually screen over 100 million ZINC compounds and identified 120 with significant growth inhibition against Escherichia coli uncovering eight structurally distinct antibiotics. Similarly, Machine learning (ML) algorithms have played a pivotal role in identifying novel inhibitors for beta-secretase (BACE1), a key enzyme in Alzheimer's disease progression~\cite{dhamodharan2022machine}. AI-driven methodologies are also expediting drug discovery for other conditions, such as SARS-COV-2~\cite{bung2021novo} and nervous system diseases~\cite{vatansever2021artificial}. These advances promise to overcome longstanding challenges in pharmaceutical research and facilitate the creation of better medications.
\subsection{Importance of MPP}
 In pharmaceutical research, predicting molecular properties is crucial for identifying viable drug candidates with desirable pharmacokinetic, pharmacodynamic, and safety profiles. Essential experimental approaches such high throughput screening, assay formulation, and toxicological studies are employed in the drug development process. However, despite significant efforts, only one out of every five compounds that enter clinical trials ultimately receive market authorization. The financial investment required further, complicates this process. With an estimated average cost of \$2.8 billion, experimentally testing billions of compounds for their suitability is both time-consuming and financially burdensome~\cite{hecht2009computational}. Therefore, accurately predicting properties such as bioactivity, solubility, permeability, and toxicity allows researchers to prioritize compounds for further experimental validation.
Computer-assisted approaches, like QSAR, offer rapid prediction of these properties through mathematical models. These methods link molecular structures to biological processes, allowing rapid molecule profiling and identifying candidates for further screening, design, and optimization~\cite{cherkasov2014qsar}. By focusing on compounds with favorable properties, researchers can streamline the drug discovery process and allocate resources more efficiently. 

\subsection{Previous Surveys and Comparative Studies}
\begin{figure}[ht!] 
 \centering
 \includegraphics[scale=0.077]{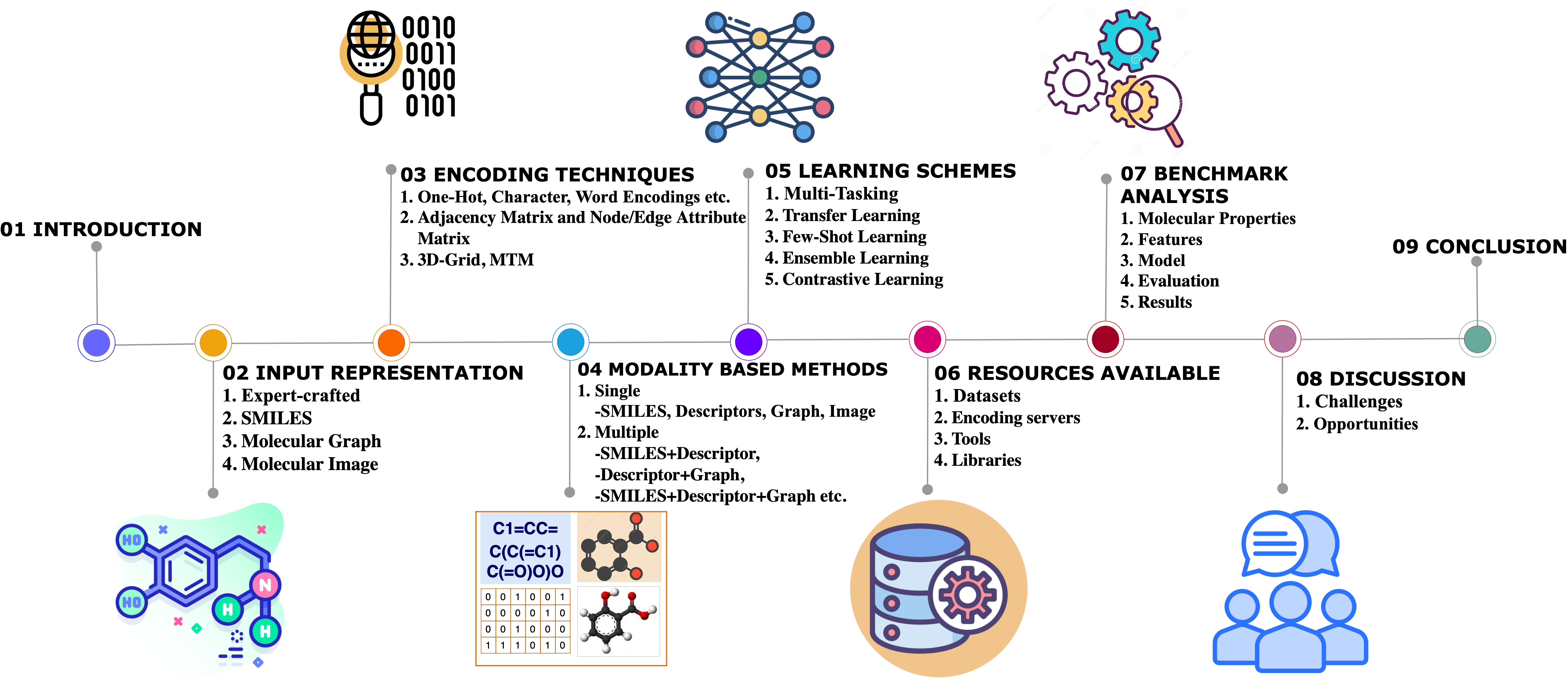}
  \caption{The structure of the overall review.}
  \label{fig:taxonomynew}
\end{figure}

This section provides a comparative analysis between our survey and previous surveys. While many previous surveys have focused on specific aspects of MPP, our survey takes a comprehensive approach and covers various dimensions of the field. For instance, Shen and Nicolaous~\cite{shen2019molecular} delve into the description of ML techniques utilized in property predictions and the diverse representations employed for this purpose. Walters and Barzilay~\cite{walters2020applications} summarize neural network-based approaches and ML methods along with their associated molecular representations. Wieder et al.~\cite{wieder2020compact} present an overview of various GNNs and their variants applied to the prediction of one or more molecular properties, showcasing the versatility of GNNs in these tasks. Li et al.~\cite{li2022deep} offer a comprehensive overview of DL techniques across various molecule data formats which include 1D, 2D, and 3D representations. Guo et al.~\cite{guo2022graph} focus on graph-based molecule representation learning and its utility in property prediction in related domains such as molecule production, reaction prediction, and drug-drug interaction analysis. Hu et al.~\cite{hu2023deep} survey DL applications in drug discovery including property prediction.\\
Existing reviews highlight recent trends in MPP but lack comprehensive discussion and analysis of multimodal-based methods for property prediction. To address this gap, we present a taxonomy that examine both single and multimodal-based methods, alongside prevalent learning schemes. By categorizing and analyzing these approaches, we aim to provide readers with a detailed insight into the various techniques used in MPP. The structure of the overall overview is outlined in Figure ~\ref{fig:taxonomynew}. Additionally, to offer insights into the unique contributions of each study, a comparison of various aspects of MPP covered by different reviews including ours is presented in Table~\ref{review_table}. Overall, the significant contributions made by this survey are summarized as follows:
\begin{enumerate}
    \item \textbf{Encoding Scheme Review.} It explore the diverse representations available for molecules as shown in Figure~\ref{fig:inputexpression} 
    Furthermore, it offer a concise discussion on the encoding schemes used to transform raw data, such as SMILES and molecular structure to provide insights into the preprocessing steps crucial for model input. The illustration on encoding schemes is given in Figure \ref{fig:encodings}.
    
    \item \textbf{Modality-based Taxonomy.} A taxonomy for modality-based MPP, is outline as illustrated in Figure~\ref{fig:taxonomy}, encompassing methods from both single and multiple modalities-based approaches. 
    \item \textbf{Architectural and Training Strategies.}  The survey delves into
    the decision points involved in their construction and training of standard DL models and present prevalent learning schemes utilized by researchers for enhancing MPP performance.
    \item \textbf{Datasets and Tools.}  Popular benchmark datasets and the availability of potential tools and services for feature generation is provided. Additionally, the paper present the top methods for each benchmark dataset in literature to offer insights into the efficacy of various predictive models.
    \item \textbf{Challenges and Future Directions.} A discussion on the pressing challenges is presented highlighting both the existing opportunities and areas that require further exploration.
\end{enumerate}
In summary, Section~\ref{sec:availability_DME} covers various input representations for molecular depiction. Section~\ref{sec: encoding_schemes} provides a succinct overview of encoding schemes for SMILES and Graph molecular data. Section~\ref{sec:modality_based_MPP} explores both single and multimodal methodologies in MPP. Section~\ref{sec:nn_ls}, discuss standard DL models and learning schemes for property prediction.  Section~\ref{sec:resource_availability} examines MPP datasets, tools and servers, and include performance metrics of the top-5 methods reported in the literature across different molecular properties. Section~\ref{sec:CA and EP} provides benchmark analysis of the methods and models involved in MPP. Section~\ref{sec:discussion} addresses challenges, opportunities  and future research trends in the field. Finally, Section~\ref{sec:conclusion} conclude the paper.
\begin{table}[htbp]
    \centering
    \caption{Comparison of the survey with others on MPP in terms of Molecule Input Expression (MIE), Encoding Techniques (ET), Modality Methods (MM), Learning Schemes (LS), Resources (RE), and Benchmark Analysis (BA).}
    \label{review_table}
    \scalebox{0.60}{
    \begin{tabular}{ p{1.5cm} p{3cm} p{3cm} p{3cm} p{3cm} p{3cm} p{3.2cm} p{1.0cm} }
 \hline
 \textbf{Article} & \textbf{MIE}  &  \textbf{ET} &
\multicolumn{2}{|c|}{\textbf{MM}} &  \textbf{LS} & \textbf{RE} & \textbf{BA} \\
 \hline
  & & & \textbf{Single} & \textbf{Multiple} & & &\\
 \hline
 Shen and Nicolaous~\cite{shen2019molecular}   & \ding{55} & \ding{55} & Descriptors,Fingerprint, \newline SMILES,Graph & \ding{55} & Multitask Learning,\newline Transfer Learning & \ding{55} & \ding{55}\\
 Oliver et al.~\cite{wieder2020compact} & \ding{55} & \ding{55} & Graph & \ding{55} & Supervised,Unsupervised,\newline Semi-Supervised, \newline Reinforcement Learning & \ding{55} & \ding{55}\\ 
 Li et al.~\cite{li2022deep} & 1D,2D,3D & \ding{55} & SMILE,Graph,Image & SMILES+Graph, \newline Descriptors+SMILES+\newline Graph & Transfer Learning, \newline Meta-Learning, \newline Multi-Task Learning & \ding{55} & \ding{55}\\
 Hu et al.~\cite{hu2023deep} & SMILES,Fingerprints,\newline Graph & \ding{55} & SMILES,Fingerprints,\newline Graph &  \ding{55} & \ding{55} &  Databases,\newline Datasets & \ding{51}\\
 Ours & Fingerprints,Descriptors,\newline SMILES,Image,Graph & SMILES encodings,Graph encodings\newline, Image encodings & Fingerprints,Descriptors,\newline SMILES,Image,Graph & SMILES+Graph,\newline Fingerprint+Graph, \newline Fingerprint+Graph+\newline SMILES,\newline SMILES+Graph+Image & Multi-task learning,\newline Transfer Learning,\newline Few-shot Learning,\newline Ensemble Learning & Datasets,\newline Encoding Servers,Tools \newline,Libraries & \ding{51}\\
\hline
\end{tabular}}
\end{table}

%% file: availability_DME.tex
\section{Availability of Different Molecule Input Expressions}\label{sec:availability_DME}
Researchers and chemists have several input expressions to facilitate the investigation, analysis, and prediction of molecular properties. This section provides a concise overview of the commonly used molecule input expression for property prediction.
\begin{figure}[ht!] 
 \centering
 \includegraphics[scale=0.35]
{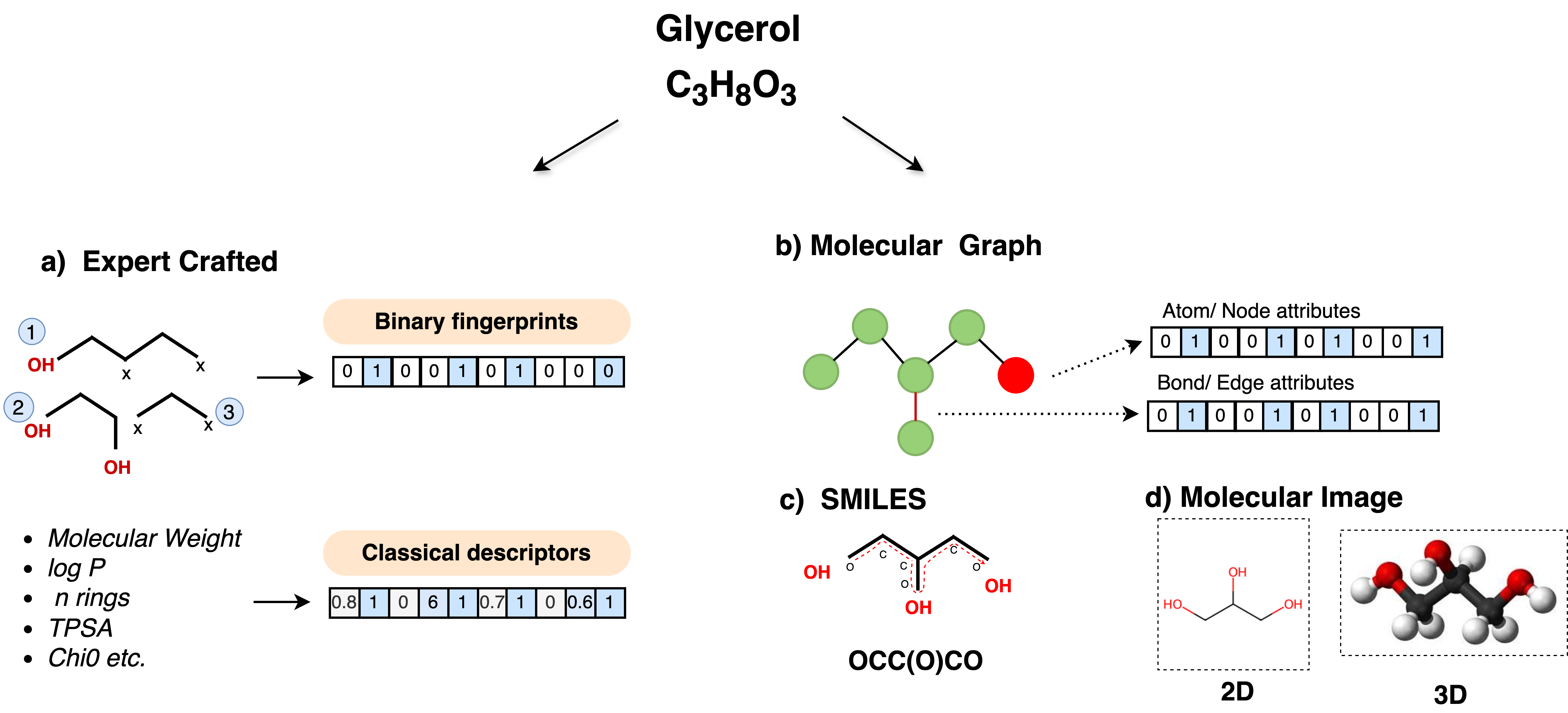}
  \caption{Various input representations of molecules utilized in MPP}
  \label{fig:inputexpression}
\end{figure}
\subsection{Expert-crafted Features}
The term "expert-crafted features" refers to chemical descriptors and fingerprints manually developed by experts in chemistry, bioinformatics, and related fields. These features encapsulate molecular traits, structural characteristics, and domain-specific information of chemical compounds, effectively capturing crucial molecular properties and structural patterns~\cite{xue2000molecular}. 
\begin{itemize}
    \item \textbf{Molecular descriptors.} These numerical representation of chemical properties are fundamental in QSPR modeling. Various types of descriptors capture different facets of molecular structure and properties ~\cite{khan2016descriptors}. For example, topological descriptors reflect molecular structure through connectivity patterns, while electronic and geometrical descriptors detail electronic properties and molecular geometry. Constitutional descriptors outline molecular composition and atom connectivity. Physicochemical descriptors, 3D descriptors, among others, further enrich the molecule representation. By integrating these descriptors, researchers develop robust models for accurately predicting molecular activities across diverse chemical space~\cite{grisoni2018impact}. 
    \item \textbf{Molecular fingerprints.} Molecular fingerprints are binary bit strings that represents the substructural features of molecules. Two widely used methods for molecular fingerprinting are key-based fingerprints and hash fingerprints. Key-based fingerprints, examplified by molecular access system (MACCS)~\cite{durant2002reoptimization} and the PubChem fingerprint~\cite{pubchem}, employ a predetermined fragment library to encode molecule into a binary vector based on its substructures. MACCS utilizes 166 predefined fragments and is compatible with cheminformatics software tools like RDKit~\cite{rdkit}, and CDK~\cite{cdk}. Conversely, the PubChem fingerprint has 881 bits representing features such as element count, ring system type, atom pairing, nearest neighbors, detailed information is available in the corresponding document~\cite{pubchem}. Hash fingerprints, unlike key-based fingerprints, generate unique binary representations for molecules using hashing algorithms. These algorithms hash the molecular structure into a unique identifier, which is then converted into a binary fingerprint. Common hash fingerprint algorithms include daylight, extended-connectivity fingerprints (ECFP), and topological torsion fingerprints (TTFP).
    \end{itemize}
\subsection{SMILES}
Introduced by Weininger~\cite{weininger1988smiles} in 1987, SMILES provides a concise notation for representing molecular structures. SMILES encode essential structural information such as atom types, bonds, connectivity, and stereochemistry using a sequence of characters and symbols. However, they do come with limitations. The length of SMILES representation varies with the size of the molecule, presenting challenges in developing generic models. Additionally, SMILES lacks internal canonicalization, allowing atoms to be mapped in any order that results in multiple notations for a single compound. Despite these limitations, SMILES strings are widely used in cheminformatics, particularly as inputs for natural language processing (NLP) algorithms~\cite{parakkal2022deepbbbp, yang2022ensemble}, which have shown promising results~\cite{honda2019smiles}. 
\subsection{Molecular Graph}
Molecular graphs serve as highly detailed representations of molecules. Within these graphs, nodes typically represent individual atoms with features such as atomic number, hybridization state, and other properties. Edges between nodes signify the chemical bonds between atoms, encoding crucial details such as bond type (e.g., single, double, or triple bonds) and bond length. DL methods, particularly GNNs, has shown promising results in tasks such as property prediction by effectively extracting the complex relationships encoded within molecular graphs~\cite{coley2017convolutional}.
\subsection{Molecular Image}
Molecular images~\cite{yoshimori2021prediction}, often denoted as 2D or 3D visual representations of molecular structures play a crucial role in property prediction methodologies as it allows researchers to visually inspect and analyze the structure of molecules. These images serve as inputs for models tasked with predicting molecular properties. The visual nature of these representations enhances interpretability and allow researchers to discern how changes in molecular structure correlate with changes in predicted properties~\cite{zeng2022accurate}.

%% file: encoding_schemes.tex
\section{Encoding Techniques}\label{sec: encoding_schemes}
In this section, encoding schemes are categorized into three groups based on modality: SMILES encoding, molecular graph encoding, and image encoding techniques. These methods transform raw data into model-compatible representations for subsequent processing.
\subsection{SMILES Encoding Techniques}
One-hot encoding~\cite{seger2018investigation} is a prevalent technique used to convert categorical data into a format suitable for ML models. Each character (atom or bond symbol) in the SMILES string is mapped to a unique index and represented as a binary vector with the value corresponding to the index of the character set to 1 and all other values set to 0. This is known as character-level tokenization, where each character in a sequence is treated as a separate token, and then encoded for model input. Word-level tokenization, on the other hand, tokenizes the SMILES string into individual words or chemical fragments, assigning a numerical index to each unique word or fragment. This is similar to one-hot encoding but uses words instead of characters. For example, given the SMILES string ‘CC(=O)O=C’, the following examples illustrate one-hot encoding, character-level tokenization, and word-level tokenization.
\begin{itemize}
    \item \textbf{One-hot encoding.} first step is to identify the unique characters present, which in this case are [‘C’, ‘(’, ‘=’, ‘O’, ‘)’]. These characters are then one-hot encoded. For instance, the one-hot encoding of ‘C’ is [1, 0, 0, 0, 0], ‘(’ is [0, 1, 0, 0, 0]. Likewise, the vector is generated for all unique characters resulting in a matrix with dimensions $n \times m$, where $n$ is the length of the SMILES string and $m$ is the number of unique characters identified.
    \item \textbf{Character-level tokenization.} SMILES strings are tokenized into individual characters. For example, the SMILES string ‘CC(=O)O=C’ may be tokenized into the characters [‘C’, ‘C’, ‘(’, ‘=’, ‘O’, ‘)’, ‘O’, ‘=’, ‘C’].
    \item \textbf{Word-level tokenization.} Word-level tokenization of the given SMILES string may result in the following tokens: [‘C’, ‘(’, ‘=O’, ‘)O’]. Each token represents a meaningful unit in the SMILES string, such as individual atoms (‘C’), functional groups (‘=O’), and parentheses (‘(’, ‘)’) used to denote branching or cyclic structures.
\end{itemize}

One-hot encoding provides a straightforward representation by encoding each character in the string while preserving the sequence. However, it results in sparse vectors that may not capture intricate relationships or important connections between characters and substructures, especially with large strings. Therefore, researchers investigate more sophisticated encoding strategies depending on the task and the properties of the data. Word-level tokenization, for instance, can capture semantic and substructure information. For example, tokenizing ‘C(=O)O’ into [‘C’, ‘(’, ‘=O’, ‘)’, ‘O’] provides insights into atoms arrangement and the presence of a carbonyl group. Word2vec~\cite{church2017word2vec} is a widely adopted method for generating word representations in a continuous vector space. For SMILES encoding, each character or token in a SMILES string is treated as a 'word', with vector embeddings learned from co-occurrence patterns. Building on Word2Vec, methods like SPVec~\cite{zhang2020spvec} have emerged. 
 SMILES pair encoding~\cite{li2021smiles}, unlike character-level tokenization, operates on substructures, to capture more detailed structural features. SMILES2Vec~\cite{goh2017smiles2vec} encodes entire strings into fixed-length vectors using RNNs, processing strings character by character. The effectiveness of these techniques depends on the granularity chosen for tokenization, and semantic coherence. Byte Pair Encoding (BPE) iteratively merges the most frequently occurring pairs of characters into a subword vocabulary that represents meaningful units in chemical structures~\cite{shibata1999byte}.
Skip-gram~\cite{preethi2020word}, a similar embedding technique, transforms SMILES into numerical vectors by predicting tokens likely to appear nearby a given target token, capturing relationships and similarities between different parts of the chemical structure that are based on context. These context-based embeddings improve model ability to understand and interpret molecular properties for various prediction tasks.
\subsection{Graph Encodings}
Graph encodings are mathematical representations of molecular structures that enables computational analysis. These encodings typically involve three matrices: adjacency matrix, node attribute matrix, and edge attribute matrix.
\begin{itemize}
    \item \textbf{Adjacency matrix.} The adjacency matrix captures the connectivity between atoms within a molecule. It is a square matrix where each row and column corresponds to an atom. The entries indicate the presence (1) or absence (0) of bonds between pairs of atoms. Additionally, the node attribute matrix and edge attribute matrix encode further information about the nodes (atoms) and edges (bonds) of the molecule.
    \item \textbf{Node attribute matrix.} The node attribute matrix contains information about the attributes or properties associated with each node (atom) in the molecular graph. Each row corresponds to a node, and each column represents a specific attribute or feature, such as atom type (e.g., carbon, hydrogen, oxygen), atomic number, mass, charge, hybridization state, and other chemical properties. Encoding these attributes in a matrix format facilitates the incorporation of node-specific information into graph-based ML models.
    \item \textbf{Edge attribute matrix.} The edge attribute matrix stores details about the properties of each bond within the molecular graph. Rows represent individual edges connecting pairs of nodes, while columns correspond to specific attributes or features of these edges. Attributes include bond type (single, double, triple), bond length, bond angle, torsion angle, and other geometric or chemical characteristics relevant to atom interactions. Incorporating edge-specific information enhances machine learning models' ability to discern intricate chemical interactions and structural patterns within molecular graphs.
\end{itemize}

\begin{figure}[ht!] 
\centering
 \includegraphics[scale = 0.40]{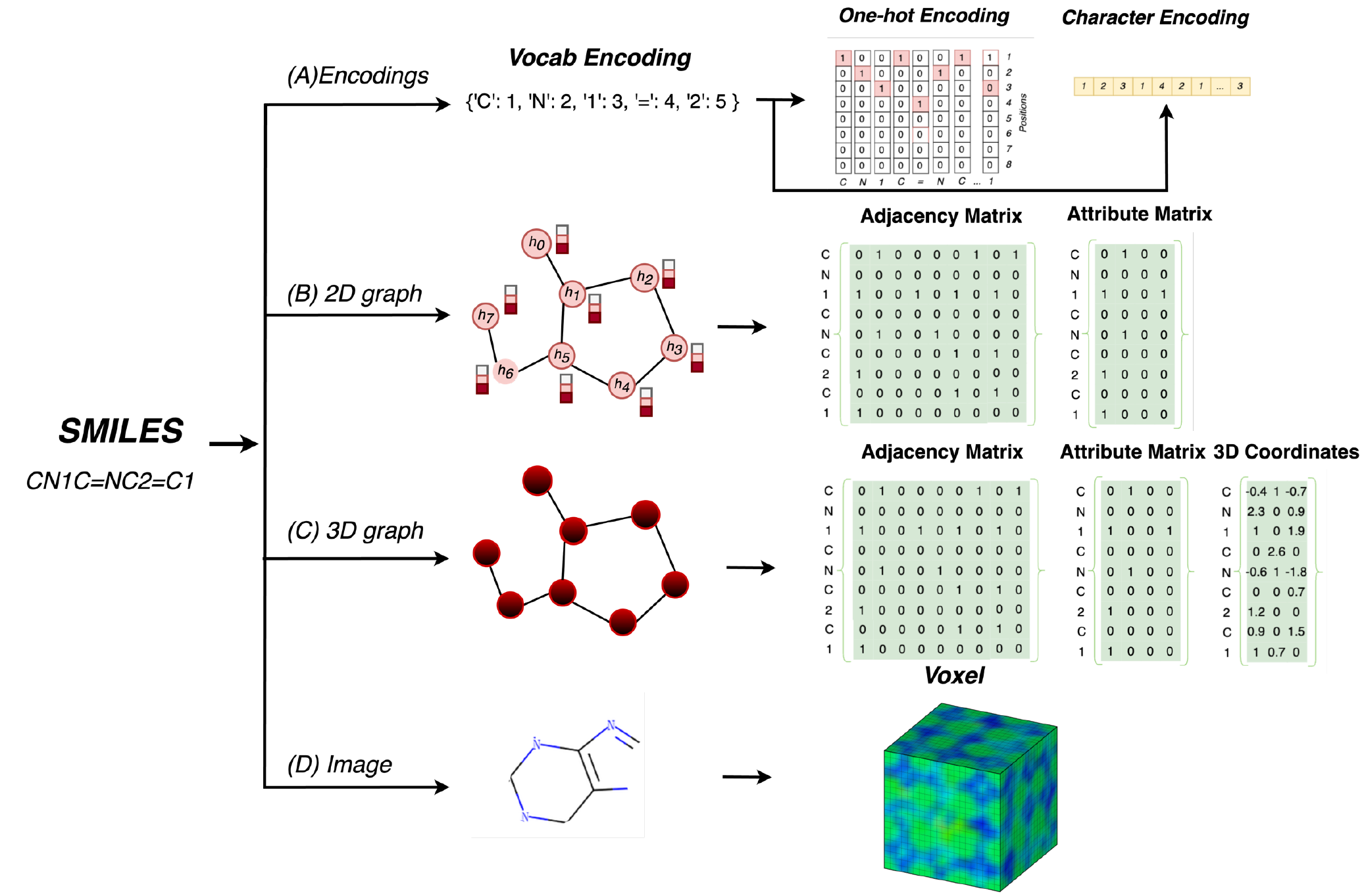}
  \caption{Encoding methods used for encoding SMILES, molecular graph and molecular images into a model processing format.}
  \label{fig:encodings}
\end{figure}
\subsection{Image Encoding techniques}
\begin{itemize}
    \item \textbf{
    Molecular Topographic Maps (MTMs).} MTMs are a representation technique that encodes molecular structures into a two-dimensional format. They are particularly useful as an image encoding method for depicting molecular structures, suitable for models like CNNs. MTMs capture structural and physicochemical attributes of molecules, including atom distribution, functional groups, and electrostatic potentials. They employ colors or contour lines to delineate different regions of the molecule, visually portraying spatial distribution and features. For example, specific colors may indicate the presence of particular functional groups or electron density levels. These images serve as input data for image-based property prediction models.
    \item \textbf{3D grid.} The 3D grid representation is a notable method within MPP categorized under image-encoding techniques. This approach effectively captures spatial information inherent in molecular structures by transforming them into a grid of voxels, akin to pixels in a 2D image. Each voxel in the grid corresponds to a localized region of space surrounding the molecule, incorporating features such as atomic charges and molecular density. The resulting grid can be visualized as a 2D image or treated as a 3D tensor. Similar to conventional images, machine learning models, particularly CNNs adept at processing visual data, can analyze patterns and relationships within the molecular data encoded in these 3D grids. Integrating CNNs with 3D grid representations enables researchers to gain deeper insights and make accurate predictions across various property prediction tasks.
    \end{itemize} 

%% file: modality_based_MPP.tex
\section{Modality-based MPP}\label{sec:modality_based_MPP}
\begin{figure}[ht!] 
 \centering
 \includegraphics[scale=0.55]{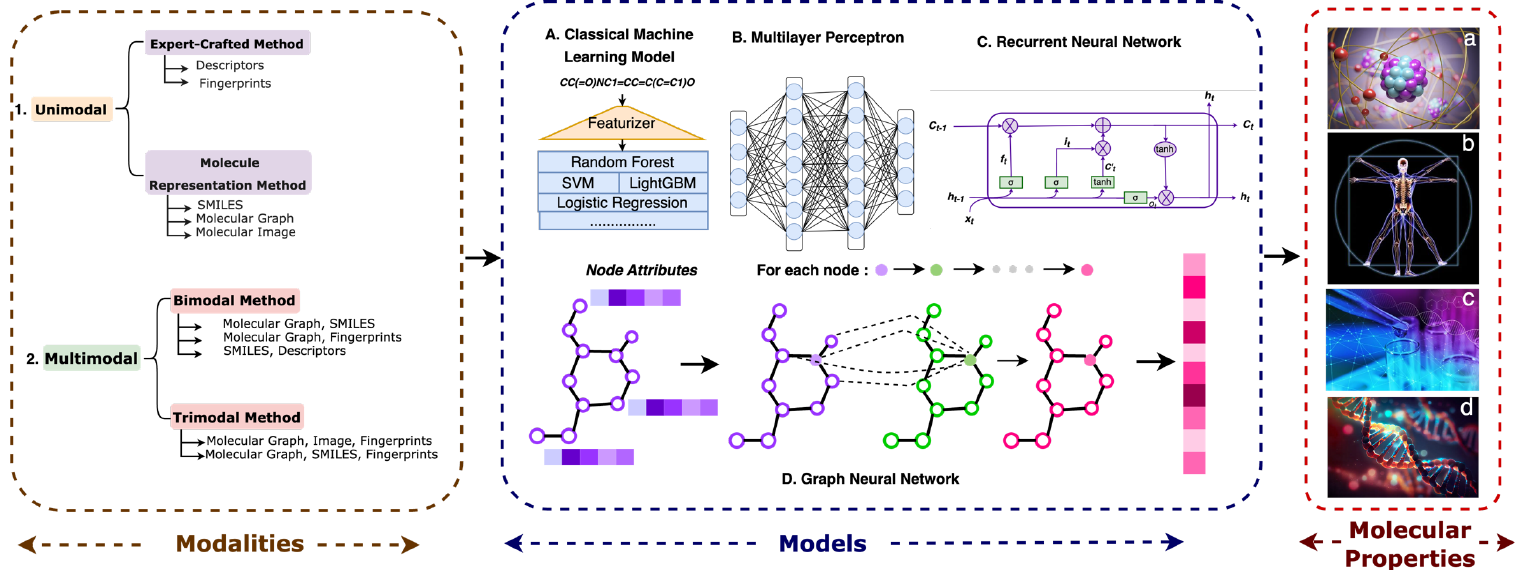}
  \caption{Modality based taxonomy of various molecular property areas including (a) Quantum chemistry (b) Physiological (c) Physical chemistry (d) Biophysics}
  \label{fig:taxonomy}
\end{figure}

This section reviews the evolution of modality-based MPP techniques, progressing from traditional handcrafted features to raw chemical data. Initially, MPP relied on manually crafted features 
using conventional ML models. The advent of DL enabled direct utilization of raw chemical compound data.
We explore methods based on single and multiple modalities. Figure~\ref{fig:taxonomy} illustrates a modality-based taxonomy for MPP. Single modality techniques consist of methods using either expert-features, SMILES, graph, or image modality. Multiple modality techniques include methods that integrate diverse representations like hand-crafted features, graph structures, and images to enhance prediction accuracy.
The input of each category are fed as input to ML and DL architecture for predicting molecular properties.

\subsection{Single modality techniques}
\subsubsection{Expert-crafted feature based methods}
 Expert-crafted feature-based methods are integral to MPP. Table~\ref{tab:leader} provides a comprehensive overview of various ML techniques used in these methods. For domain-specific information, the descriptors are frequently used as input to ML algorithms to predict molecular properties. 
 Mucs~\cite{zhang2019lightgbm} uses LightGBM to predict toxicity against various endpoints using the Tox21 and mutagenicity dataset and compared it to XGBoost, deep neural networks (DNN), random forest (RF), and support vector machine (SVM). Using 12 molecular fingerprints of 1003 structurally different compounds, Zhang ~\cite{zhang2017carcinopred} presented three unique ensemble models based on RF, SVM, and XGBoost to predict the carcinogenicity of drugs. Researches~\cite{zhang2017carcinopred, fu2019systematic} show that testing multiple descriptors, algorithms, and hyperparameters on diverse data splits is essential for creating stable, high-performance models. Sheridan~\cite{sheridan2016extreme} evaluated various ML models across 30 datasets with diverse on-target and off-target properties. The datasets such as BACE and hERG were assessed for computational time, space utilization, and performance metrics. Ensemble models outperformed individual ML models and rule-based systems, especially in sensitivity. Yang~\cite{yang2022ensemble} introduced SPOC, a molecular representation combining fingerprints with commonly used descriptors in QSAR/QSPR. Tested on 12 datasets
 SPOC showed significant potential. The combination of Avalon and atom-pair fingerprints with RDKit descriptors achieved the highest performance across all tasks.\\
 The effectiveness of predictive models often depends on the quality of expert features, though these may not always be optimal. With the rise of SMILES, molecular graphs, and molecular images, the benefits of traditional feature selection have diminished, blurring the lines between handcrafted and machine-learned features. In conclusion, while traditional descriptors and fingerprints have been essential in MPP, the field is shifting towards machine-learned representations. This transition underscores the importance of evaluating and refining modeling techniques to adapt to new methodologies and enhance predictive performance in drug discovery and related fields.
\begin{table*}[ht!]
    \begin{center}
    \caption{An overview of expert-crafted feature based approaches for MPP}
    \scalebox{0.8}{
    \begin{tabular}{c p{1.4cm} p{1.8cm} p{3.5cm} p{3.5cm}  p{1.8cm} c}\hline
    \label{tab:leader}
    \textbf{Year} & \textbf{Dataset} & \textbf{Task} & \textbf{Input\newline Representation} & \textbf{Method} &  \textbf{Evaluation criteria} & \textbf{Reference}\\
    \hline
    \multirow{1}{*}{2017} & AMES &  Classification & Descriptors, ECFP-14 & Naïve Bayes & 5-Fold CV & Zhang et al.\cite{zhang2017novel}\\
    
     
    \hline
    \multirow{4}{*}{2018} & eChemPortal
     & Classification & Descriptors,PubChem, MACCS,Substructure, CDK,Estate & SVM,KNN,Naive Bayes,DT,RF,ANN & 5-Fold CV  & Fan et al.\cite{fan2018silico}\\
    
     & BBBP & Classification & Descriptors,PubChem, Klekota-Roth,CDK-Extended,2D-Atom Path,FP4 & SVM & Train-Test Split & Yuan et al.\cite{yuan2018improved}\\
     
    & BBBP & Classification & Descriptors,Fingerprints & LR,RF,KNN,SVM,MLP & 10-Fold CV &  Wang et al.\cite{wang2018silico}\\
    
    &  ADME & Classification, \newline Regression & ECPF-3 & DNN, SVM & 10-Fold CV & Zhou et al.\cite{zhou2018exploring}\\
     
    \hline
     \multirow{1}{*}{2019}
    & $Log D_{7.4}$ & Classification & Multiple Descriptors & Consensus of RF, XGboost,SVM,GB  & Random Split  & Fu et al.\cite{fu2019systematic}\\
    
    \hline
    
    \multirow{1}{*}{2021} & OECD-TG471 & Classification & Fingerprints & Balancing Techniques,GBT,RF, SVM, MLP,KNN & Random Split &  Bae et al.\cite{bae2021effective}\\
    \hline
    \multirow{2}{*}{2022} & BBBP & Classification & Mol2Vec &  1D-CNN, MLP & 10-Fold CV & Parakka et al.\cite{parakkal2022deepbbbp}\\
    & HIV,BACE,\newline QM7,Lipo,\newline BBBP,ESOL etc. & Classification,\newline Regression & Descriptors,Fingerprints  & RF & Random Split & Yang et al.\cite{yang2022ensemble}\\
    
\hline
\end{tabular}}
\end{center}
\end{table*}
\subsubsection{{SMILES}}
SMILES serves as a widely adopted format for representing chemical compounds in various databases \cite{weininger1988smiles}. 
Recent advancements in NLP have enabled effective integration of SMILES sequences into property prediction tasks. By employing preprocessing techniques and learning schemes, these methods extract meaningful features from SMILES to optimize predictive modeling. Similar to image analysis techniques that use transformations like blurring or rotation to expand training datasets, SMILES-based approaches utilize augmentation methods to enhance data for better representation learning~\cite{shorten2019survey}. For example, Kimber et al.~\cite{kimber2021maxsmi} employed five SMILES augmentation methods across physicochemical datasets and showed improvements in terms of accuracy. Chen and Tseng~\cite{chen2021different} demonstrated the effectiveness of CNNs using the convS2S architecture. Augmented SMILES strings are translated into embedding vectors through a count-based dictionary. An encoder network, comprising a convolutional block, gated linear units (GLUs), and fully connected layers, processes these vectors. Positional embeddings are also incorporated to capture sequential information. SCFP, as proposed by Hirohara et al.~\cite{hirohara2018convolutional}, also uses CNN that processes one-hot encoded SMILES vectors. CNN-based models, though effective, necessitate a fixed input sample length, requiring drug SMILES to be padded or truncated prior to be fed as input. This is approached in two ways, by taking either the average of SMILES length or maximum SMILES length. However, both these methods cause either data loss or introduce noise.
In addition to CNNs, RNNs, particularly LSTM and GRU, have gained widespread adoption for sequence processing. Li et al.~\cite{li2022novel} introduced a hybrid architecture that comprises of stacked CNN and RNN layers for representation learning. Segler et al.~\cite{segler2018generating} demonstrated that an RNN trained on molecular SMILES strings effectively capture syntactic structure in SMILES and the distribution of chemical space. Similarly, Huo et al.~\cite{hou2022accurate} exploited SMILES using a Bidirectional Long Short-Term Memory (BiLSTM) network augmented with channel and spatial attention modules.  While RNNs, when combined with augmentation techniques, demonstrate proficiency in capturing sequence information, they may not adequately encapsulate atomic relations and bond types, which are vital for understanding molecular structures. Furthermore, while enumeration techniques facilitate the creation of larger datasets, molecules are not conducive to such implementations. Even a minor alteration in the position of a single atom within a molecule can profoundly influence its biological activity. To address these limitations, researchers have explored methods to incorporate functional groups information, as SMILES sequences may lack direct provision for such details. Contextualized architecture named Mol2Context-vec~\cite{lv2021mol2context}, employs a BiLSTM framework that enables the integration of diverse internal states to generate dynamic representations of molecular substructures. The resulting molecular representation facilitates the capturing of interactions among atomic groups, particularly those that are spatially distant. Shao et al.~\cite{shao2022s2dv} employed a optimized word2vec model for accurately representing the relationship between a compound and its substructure. The model demonstrated effectiveness in predicting the inhibitory effect of compounds on HBV and liver toxicity. Table~\ref{tab:leader4} provides comprehensive details of SMILES based methods which includes information on datasets used, encoding techniques employed, evaluation metrics, and other relevant parameters.\\
The SMILES and transformer-based architectures have also emerged as pivotal tool revolutionizing the field of MPP. A common approach involves adapting pre-trained transformer models, such as BERT(Bidirectional encoder representations from transformers) ~\cite{devlin2018bert} for the extraction of atomic or molecular attributes from SMILES sequences ~\cite{wang2019smiles}. Besides BERT, generative methods also exist for molecular representation learning that employs encoder-decoder architectures. Hu et al.~\cite{hu2020deep} employed a Gated Recurrent Unit (GRU) based encoder-decoder model to generate fixed-dimensional latent features representing molecules from SMILES strings, subsequently employing a CNN model for downstream tasks. Transformers can also be customized for MPP by designing task-specific architectures, such as transformer based encoders or decoders. These architectures may include additional layers or modules tailored to handle molecular data to capture structural dependencies in molecular graphs. For instance, the transformer model primarily designed for sequence-to-sequence tasks has been adapted for both discriminative and generative purposes by leveraging its encoder and decoder subunits independently. Molecular transformer models, such as SMILES-Transformers (ST)~\cite{honda2019smiles}, ChemFormer~\cite{irwin2022chemformer}, and Transformer-CNN~\cite{karpov2020transformer}, utilize variants of the transformer model like BERT~\cite{devlin2018bert}, BART~\cite{lewis2019bart}, and RoBERT~\cite{liu2019roberta} as base models. ST employs the transformer model and uses the encoder output as a molecular embedding for MPP. ChemFormer adopts the BART model, utilizing the transformer as a denoising autoencoder and leveraging multiple SMILES representations for data augmentation. Similarly, Transformer-CNN is trained to produce different valid SMILES representations for the same molecule. While some studies focus solely on the encoder model for discriminative tasks, others like SMILESBERT~\cite{wang2019smiles} and MolBERT~\cite{li2021mol}, employ the encoder with pre-training objectives tailored to molecular properties and chemical language comprehension. These objectives aims to predict various physico-chemical properties and enhances the model understanding of SMILES non-uniqueness by identifying equivalent representations. 
Transformers also extend to capture geometric information, a crucial aspect of molecular structures~\cite{choukroun2021geometric, kwak2021geometry}. Transformers with their ability to attend to global and local dependencies in data sequences, offer a promising approach to effectively encode and understand the geometric arrangements of atoms within molecules. 
\begin{table*}[ht!]
    \begin{center}
    \caption{An overview of the SMILES based methods developed for MPP}
    \scalebox{0.64}{
    \begin{tabular}{c p{3.0cm} p{2cm} p{2cm} p{2.5cm} p{3.0cm} p{1.5cm} p{3cm} p{2cm}}
    \hline
    \label{tab:leader4}
    \textbf{Year} & \textbf{Dataset} & \textbf{Task} & \textbf{Input} & \textbf{Encoding} & \textbf{Method} &  \textbf{Evaluation} &  \textbf{GitHub/ \newline Server} & \textbf{Reference}\\
    \hline
    \multirow{3}{*}{2019}
     & BBBP,BACE,Ames,\newline ESOL& Classification,\newline Regression & SMILES,\newline InCL & Tokenization & CNN,RNN & Random CV,\newline Cluster CV &  jrwnter/cddd & Winter et al.~\cite{winter2019learning}\\
    & PubChem &  Regression & SMILES & Skip-Gram & Tree-LSTM, BPNN & - &  - &  Su et al.~\cite{su2019architecture}\\
    \hline
    \multirow{4}{*}{2020}
       &  Lipo,FreeSolv,\newline HIV,BBBP & Classification,\newline Regression & SMILES & Tokenization & MolPMoFiT & Random Split & - & Li and Fources~\cite{li2020inductive}\\
       & Lipo,BACE,FreeSolv,\newline BP,HIV,AMES, \newline BBBP,ToxCast & Classification,\newline Regression & SMILES & One-hot Encoding & Transformer,\newline CNN & Random Split &  bigchem/transformer-cnn & Karpov et al.~\cite{karpov2020transformer}\\
      & HIV,BACE,BBBP,\newline Tox21,ClinTox,SIDER &  Classification & SMILES & Byte Pair Encoding,\newline Word2Vec & Message Passing & Train-Test Split & - & Jo et al.~\cite{jo2020message}\\
      \hline
      \multirow{6}{*}{2021} 
       &  Tox21,HIV,BBBP\newline SIDER,\newline CLINTOX & Classification,\newline Regression & SMILES & Atom-Embedding & Self-Attention,\newline CNN & Random Split & arwhirang/samtl & Lim and Lee~\cite{lim2021predicting}\\ 
      & QSAR datasets &  Regression &  SMILES & SMILES Pair \newline Encoding & AWD-LSTM & Random Split & XinhaoLi74/SmilesPE & Li and Fources~\cite{li2021smiles}\\
      & Tox21,BBBP\newline CLINTOX,\newline SIDER & Classification & SMILES & Morgan Based\newline Atom Identifier & BERT & Scaffold Split & cxfjiang/MolBERT & Li and Jiang~\cite{li2021mol}\\
      \hline
      \multirow{4}{*}{2022}
      & logS,logP,logD & Regression & SMILES & Word2Vec based\newline tokenization & BiLSTM,\newline CBAM,\newline MLP & Random Split &  SMILES-Enumeration-Datasets & Hou et al.~\cite{hou2022accurate}\\
      & Lipo,BACE,ESOL,\newline HIV,FreeSolv,& Classification,\newline Regression & SMILES \newline Augmentation & Tokenization & Stacked CNN, \newline and RNN & 5-Fold CV & -  & Li et al.~\cite{li2022novel}\\
       & HBV,HepG2 & Classification & SMILES & Tokenization,\newline Skip-Gram & - &  Random Split & NTU-MedAI/S2DV & Shao et al.~\cite{shao2022s2dv}\\
      & MoleculeNET & Classification,\newline Regression & SMILES & Morgan based ECFP & BERT,\newline CNN & Random Split & - & Wen et al.~\cite{wen2022fingerprints}\\
    \hline
    \multirow{1}{*}{2023}
     & MoleculeNET,\newline Cytotoxicity & Classification,\newline Regression & SMILES Augmentation & Tokenization,\newline &  1D-CNN, Multi-head\newline Attention & Random Split & PaccMann/toxsmi & Markert et al.~\cite{markert2020chemical}\\
 \hline
 \multirow{1}{*}{2024}
     & ESOL,FreeSolv\newline Lipo,BBBP,Clintox & Classification,\newline Regression & SMILES & Tokenization,\newline & CNN,BERT & Random Split & - & Yan et al.~\cite{yan2024insights}\\
\hline
\end{tabular}}
\end{center}
\end{table*}
\subsubsection{{Molecular Graph}}
\begin{figure}[ht!] 
\centering
\includegraphics[scale = 0.40]{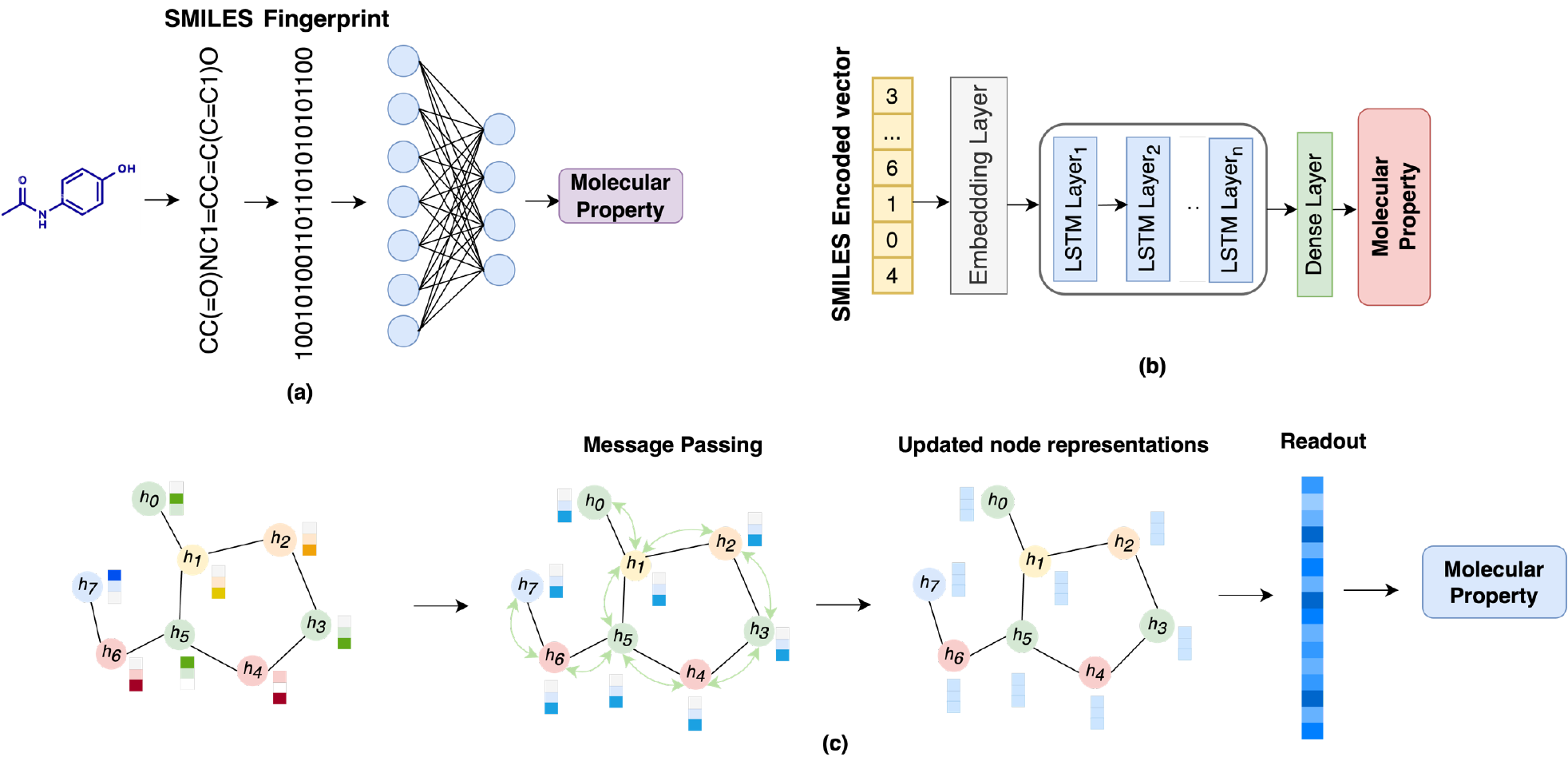}
  \caption{Illustration of MPP using (a) Descriptor-based Neural Network, (b) SMILES string-based sequential model such as LSTM, and (c) Molecular structure-based Graph Neural Network (GNN). Each approach utilizes a different input representation to predict molecular properties, showcasing the versatility of computational methods in addressing diverse challenges in drug discovery and materials science.}
\label{fig:three modalities}
\end{figure}

Graph-based methods for predicting molecular properties represent molecules as graphs, with atoms as nodes and bonds as edges. These methods leverage molecular structure and topology to extract critical information for property prediction. We provide a detail insights into graph-based methods for MPP on various parameters in Table~\ref{tab:leader2 continue}. Graph convolutional networks (GCNs) apply convolution operations over graphs to learn and propagate data across nodes and edges, capturing both local and global structural features. GCNs use two main approaches: spectral convolution and spatial convolution. Spectral convolution relies on the graph Fourier transform, which converts graph data into the frequency domain using the eigenvectors of the graph Laplacian matrix. Methods like ChebNet employ this approach. Shang et al.~\cite{shang2021multi} proposed an edge-aware spectral GCN model featuring an adaptive spectral filter. This model segmented the molecular graph into multiple views based on edge type and introduced a consistent edge-mapping mechanism to learn edge attention weights. However, spectral methods are limited compared to spatial methods due to their constraint of fixed graph sizes. Spatial convolution propagates information through the graph by aggregating features from neighboring nodes, considering their local structure. Techniques like GraphSAGE and Graph Isomorphism Networks (GINs) use spatial convolution. These models require an adjacency matrix, a node feature matrix, and an edge feature matrix. TrimNet~\cite{li2021trimnet} emphasizes edge information via a triplet-aware edge network, enhancing edge information retrieval. Gilmer et al.~\cite{gilmer2017neural} proposed a message passing neural network (MPNN) to acquire a graph-level embedding containing both node features and weighted edge messages.
Graph-based models often suffer with the issue of oversmoothing, wherein multiple layers of message passing and aggregation cause nodes to become increasingly similar to each other. To address this challenge, specialized form of graphs, such as directed graphs with directed edges, can be employed. For instance, the edge memory neural network ~\cite{tian2023predicting} focuses on edge messages, and the iteratively focused graph network (IFGN) identifies key attributes responsible for specific properties. The illustration about the graph based model along with two different modalities is shown in Figure~\ref{fig:three modalities}.\\
In many real-world applications, graphs often exhibit hierarchical organization, where entities at different levels of granularity interact with each other in complex ways. Therefore, it is crucial in GNN as well as it allow the model to capture multi-scale structures and relationships within graphs. Su et al.~\cite{su2019architecture} and Wang et al.~\cite{wang2019predictive} converted graphs into tree-based structures to capture hierarchical characteristics. Wang et al.~\cite{wang2020molecular} introduced a multi-channel tree-based method for predicting molecular structures, transforming molecules into substructure graphs traversed using the breadth-first search (BFS) algorithm. This method extracts molecular features at both node and molecule levels, capturing fine-grained and coarse-grained information. However, converting graphs to trees introduces non-uniqueness due to variations in traversal methods and root atom selection, impacting model generalizability.\\
GNNs often require large amounts of labeled data for training, which may be expensive or impractical to obtain. SSL allows GNNs to leverage large amounts of unlabeled data reducing the need for labeled examples. 
The efficacy of contrastive learning in predicting molecular properties depends significantly on selecting and generating relevant pairs of molecular structures. Studies by Rong et al.~\cite{rong2020grover} and Sun et al.~\cite{sun2021mocl} highlight the importance of well-curated pairs. Pre-training methods like GROVER~\cite{rong2020grover}, MoCL~\cite{sun2021mocl}, and MGSSL~\cite{zhang2021motif} tend to outperform conventional methods, allowing pre-trained GNN models to transfer to downstream tasks with limited labeled data. The learned representations from self-supervised tasks often capture generic structural properties of graphs, making them useful for a wide range of tasks without the need for task-specific labeled data. Spatial arrangement and orientation of atoms in a molecule, also significantly influence its properties and behavior. This conformational information significantly impacts geometric characteristics such as bond angles, dihedral angles, and bond lengths. Therefore, integrating conformational information into molecular representations allows for a more accurate depiction of a molecule's shape and geometry, enhancing MPP precision. Lu et al.~\cite{lu2019molecular} introduced a multilevel GCN that preserves both conformational and spatial information. On a similar note, Liu et al.~\cite{liu2021spherical}  proposed spherical message passing and SphereNet for 3D molecular learning, achieving generalized predictions of rotation and translation invariance. Additionally,  Liu et al.~\cite{liu2021pre} also introduced a semi-supervised method using using both 3D and 2D graphs of the same molecule to train models.
However, handling conformational flexibility and capturing relevant conformers for property prediction pose computational challenges. To address this, researchers are incorporating curvature information to enhance GNN effectiveness. Inspired by the Graphformer architecture~\cite{ying2021transformers}, Chen and Tseng~\cite{chen2023curvature} developed a curvature-based transformer model that improves graph transformer neural networks by preserving both structural and functional information with discretized Ricci curvature. The line graph transformer (LiGhT)~\cite{li2022kpgt} is another innovative transformer model specifically designed to capture the structural information of molecular graphs, emphasizing the significance of chemical bonds. These advancements signify a shift towards enhancing the structural understanding and predictive capabilities of GNN models in MPP.
\begin{table*}[ht!]
    \begin{center}
    \caption{An overview of the GNN based methods developed for MPP}
    \scalebox{0.70}{
    \begin{tabular}{c p{4.0cm} p{2cm} p{1.5cm} p{2.5cm} c p{3cm} p{1.5cm}}\hline
    \label{tab:leader2 continue} 
    \textbf{Year} & \textbf{Dataset} & \textbf{Task} & \textbf{Spatial/\newline Spectral} & \textbf{Method} &  \textbf{Evaluation} & \textbf{GitHub/ \newline Server} & \textbf{Reference}\\
    \hline
  \multirow{6}{*}{2019} & QM9, MUTAC,NCI1 & Classification, Regression & Spatial & CCN & 10-Fold CV & - & Maron et al.~\cite{maron2019provably}\\
 & ChemBL & Regression & Spatial & GraphNet & Random Split &  choderalab/gimlet & Wang et al.~\cite{wang2019graph}\\
 & QM9,COD,CSD & Regression & Recurrent & DGGNN & - & - & Mansimov et al.~\cite{mansimov2019molecular}\\
 & MUTAC & Classification  & Spatial & RGAT and RGCN & 5-Fold CV & - & Busbridge et al.~\cite{busbridge2019relational}\\
 & ESOL,LIPO,Tox21 & Classification, \newline Regression & Spatial & ExGCN & Random split & - & Meng et al.~\cite{meng2019property}\\
  & HIV,MUV,BBBP,Tox21,\newline SIDER, QM8,\newline ESOL,LIPO\newline etc. & Classification, \newline Regression & Spatial & AttentiveFP & Random Split & - & Xiong et al.~\cite{xiong2019pushing}\\
 \hline
 \multirow{6}{*}{2020}
 & HIV,MUV,BBBP,Tox21,\newline SIDER, QM8,ESOL,LIPO & Classification,\newline Regression & Spatial & AMPNN,EMNN & Random Split &   edvardlindelof/graph-neural-networks-for-drug-discovery & Withnall et al.~\cite{withnall2020building}\\
 & NCI109 & Classification &  Spatial & MxPool & 10-Fold CV &   JucatL/MxPool & Liang et al.~\cite{liang2020mxpool}\\
  & QM9,ZINC & Classification & Spatial & Local Relational Pooling & Random split & leichen2018/GNN-Substructure-Counting & Chen et al.~\cite{chen2020can}\\
 & BBBP,ADMET & Regression & Spatial & MT-PotentialNet & Temp\footnote{Temporal}+MW\footnote{Molecular Weight} Split  & - &  Feinberg et al.~\cite{feinberg2020improvement}\\
 & MACE,BBBP,Tox21,SIDER\newline ClinTox & Classification & CGNN & MVGNN & Scaffold Split & - & Ma et al.~\cite{ma2020multi}\\
\hline
\multirow{3}{*}{2021}
 & Tox21,Freesolv,Lipo,eSOL & Classification,\newline Regression  & Spectral & EAGCN & Random Split &  -    
& Shang et al.\cite{shang2021multi}\\

 \hline
 \multirow{3}{*}{2022}
& PDBbind-v2007,PDBbind-v2013,
PDBbind-v2016 & Regression & Spatial & MP-GNN & Random Split & Alibaba-DAMO-DrugAI/MGNN & Li et al.\cite{li2022multiphysical}\\
 & BACE,Tox21,QM8,QM7,\newline ESOL,Lipo,FreeSolv,SIDER & Classification,\newline Regression & Spatial & MV-GNN,\newline CD-MVGNN  & Scaffold split & uta-smile/CD-MVGNN & Ma et al.\cite{ma2022cross}\\
\hline
\multirow{3}{*}{2023}
 & Tox21,ToxCast,BBBP,BACE,\newline  ESOL,Lipo,FreeSolv,SIDER\newline  & Classification,\newline Regression & Spatial & IFGN & Scaffold split & http://graphadmet.
cn/works/IFGN & Tian et al.\cite{tian2023predicting}\\
& BBBP & Classification & Spatial & MPNN,GAT,GCN & Scaffold Split & - & Dinesh et al.\cite{dinesh2023benchmarking} \\
 & Log S & Regression & Spatial & GCN,GIN,GAT,\newline Attentive FP & 10-Fold CV & - & Ahmad et al.\cite{ahmad2023attention}\\
\hline
\multirow{1}{*}{2024}
 & ESOL,Lipo,FreeSolv,SIDER\newline Tox21,ToxCast,BBBP,BACE  & Classification,\newline Regression & Spatial & 3D-Mol & Random split & AI-HPC-Research-Team/3D-Mol & Kuang et al.\cite{kuang20243d}\\
 & ESOL,Lipo,FreeSolv,SIDER\newline Tox21,ToxCast,BBBP,BACE\newline HIV,MUV,QM7,QM8,QM9   & Classification,\newline Regression & Spatial & DIG-Mol & Random split & ZeXingZ/
DIG-Mol & Zhao et al.\cite{zhao2024contrastive}\\
& HIV,MUV,SIDER\newline Tox21,Clintox,BBBP,BACE  & Classification,\newline Regression & Spatial & AEGNN-M & Random, \newline scaffold split & Sixseven-Five/AEGNN-M) & Cai et al.\cite{cai2024aegnn}\\
\hline
\end{tabular}}
\end{center}
\end{table*}
\subsubsection{{Molecular Image}}
The success of DL in image processing has inspired advancements in MPP through image-based methods. These methods transform molecular structures into images, leveraging image analysis to capture complex molecular details and enhance prediction accuracy. Although RDKit provides tools for generating molecular images, these often lack essential visual elements like bond annotations, atom labels, and clear stereochemistry. To address this, Yoshimori introduced molecular MTMs~\cite{yoshimori2021prediction}, representing molecules as 2D matrix data. MTMs, generated using atomic features through generative topographic mapping, have demonstrated superior performance compared to traditional molecular fingerprints such as Morgan fingerprints and MACCS keys.\\
While frequency-domain methods are typically associated with signal analysis, they have also been explored in MPP to extract structural and property-related information. Converting images from the spatial to the frequency domain facilitates the separation of various image components, such as background noise, edges, and textures. Tchagang and Valde~\cite{tchagang2021time} transformed molecules into images using frequency-domain techniques by first converting the molecule to a 1D Coulomb matrix and then introducing a time-frequency-like (TFL) transformation. This approach encodes structural, geometric, energetic, electronic, and thermodynamic properties. However, image-based methodologies in MPP face challenges in transforming data samples into Euclidean space, often lacking essential atom and bond attributes necessary for precise predictions. Further research is needed to develop image generation techniques that reveal intricate relationships among atoms from specific perspectives.\\
Recent advancements in image-based DL models have contributed to MPP. One method involves using a 3D grid, which divides the space surrounding a molecule into voxels~\cite{tong2023vismole}. Processing a 3D grid typically involves 3D convolutional operations, which detect local and spatial patterns within the 3D structure. This allows the model to learn from the 3D arrangement of atoms and molecular features. Libmolgrid~\cite{sunseri2020libmolgrid} is a library for depicting 3D molecules using arrays of voxelized molecular data, supporting temporal and spatial recurrences to facilitate work with convolutional and recurrent neural networks. Although studies on 3D grids for MPP are currently limited~\cite{li2021spatial}, this approach is expected to play a significant role in advancing MPP and drug discovery efforts.
\subsection{Multiple modality techniques}
\begin{figure}[ht!] 
\centering
\includegraphics[scale = 0.35]{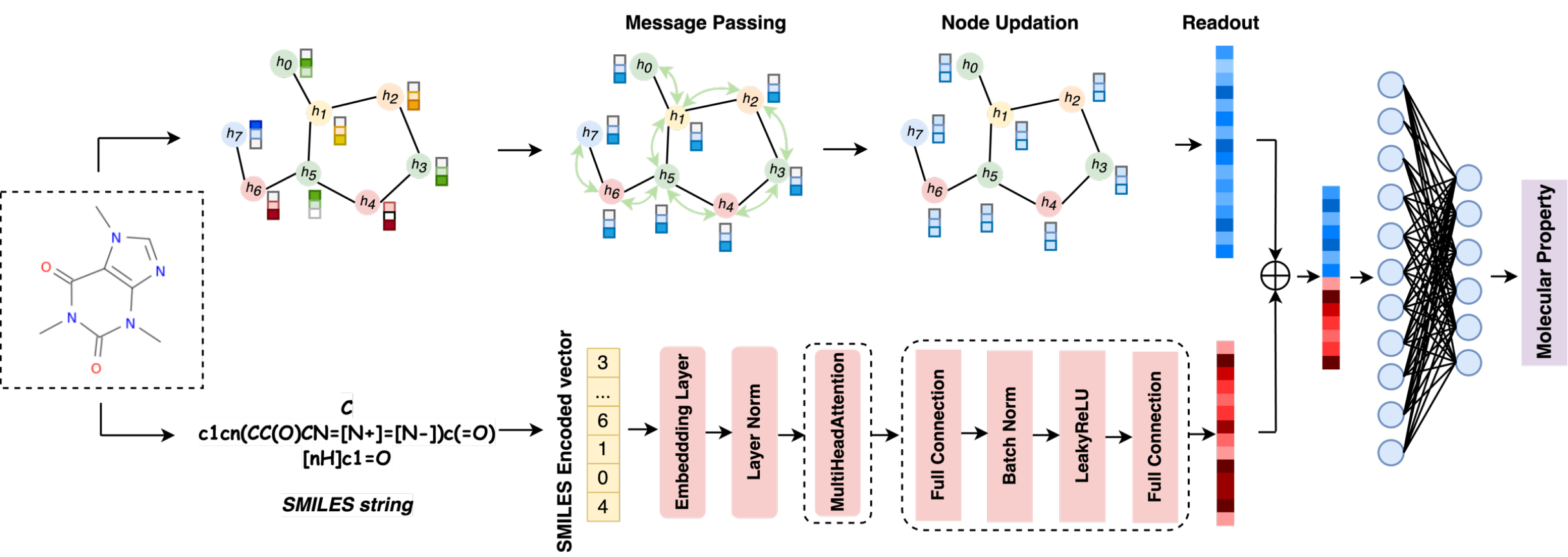}
\caption{Workflow illustration demonstrating the utilization of SMILES notation and graph structure as input data for learning molecular representations through multiple modalities in the context of MPP. This process involves encoding raw molecular data into machine-readable formats, followed by the application of representation learned to capture diverse molecular features.}
\label{fig:bimodality}
\end{figure}

Multi-modality models leverage diverse representations such as expert features, SMILES, molecular graphs, and molecular images, and integrates them to capture various essential aspects of molecular structures. We provide  the representative model and the comprehensive summary of multi-modality methods in Table~\ref{tab:hybridleader}. This comprehensive approach enhances featurization, leading to more accurate predictions. For example, GraSeq ~\cite{guo2020graseq} combines SMILES representations with molecular graphs using LSTM and GNN, capturing both sequential and topological information to enhance performance across multiple tasks. Similarly, MTBG~\cite{liu2023prediction} integrates SMILES and molecular graphs for toxicity prediction through a dual-pipeline approach, processing SMILES strings with BiGRU and learning molecular graph embeddings using GraphSAGE~\cite{hamilton2017inductive} model. We introduce a flow diagram illustrating the bimodal pipeline based on SMILES and molecular graphs in Figure~\ref{fig:bimodality}.
Incorporating chemical domain information via expert-crafted features with molecular graphs improves model performance. Wang et al.~\cite{wang2019molecule} combined GCNs with molecular fingerprints using chemopy~\cite{cao2013chemopy} that resulted in a model achieving better generalization of molecular features.
FP-GNN~\cite{cai2022fp} integrates fingerprint information with spatial information from GAT, enhancing predictive performance. In a similar approach, Dnn-PP~\cite{wiercioch2023dnn} combines molecular structure embeddings from a graph attention mechanism with DNN handled descriptors to ensure comprehensive representation of molecular features.\\ 
Integrating more than two representations has become a prevailing trend in predictive modeling, leveraging the strengths of each representation to improve overall performance. Meta-ensembling learning, as demonstrated by Karim et al.~\cite{karim2019toxicity}, uses diverse representations like molecular images, 2D features, and SMILES for toxicity prediction, integrating outputs from different neural network architectures. Another study by Karim et al.~\cite{karim2021quantitative} showcased the aggregation of base learner outputs using a meta-learner, emphasizing the significance of using multiple molecule representations. SSL approaches, such as He et al.~\cite{he2023novel}, incorporate descriptors and molecular graphs to tackle representation conflicts and training imbalances. This innovative methodology aims to tackle two primary challenges encountered in traditional encoder training: representation conflicts and training imbalances. The bidirectional encoder from transformer has also demonstrated remarkable capabilities in leveraging extensive unlabeled molecular data through SSL strategies. However, its application overlooked the crucial 3D stereochemical information inherent in molecules. To address this limitation, the use of algebraic graphs, like the element-specific multiscale weighted colored algebraic graph, incorporates complementary 3D molecular details into graph representations~\cite{chen2021algebraic}. In another approach, Busk et al.~\cite{busk2021calibrated} trained an ensemble of MPNN with random parameters on the same dataset. They enhanced model performance by calibrating MPNN results using the variance of classifiers to indicate uncertainty. Ding et al.~\cite{ding2022combining} demonstrated the impact of integrating multi-dimensional fingerprints reflecting structural and geometrical properties on hERG cardiotoxicity prediction of small molecules. Their experiments included validation on external datasets to demonstrate the efficacy of the proposed model.\\
Molecules have intricate hierarchical structural patterns, organized into functional groups and substructures. Hyperbolic graphs effectively capture these hierarchical relationships, providing a nuanced depiction of chemical structures. To enhance multi-modal systems for MPP, hyperbolic graph embeddings can be integrated with other molecular representations, such as molecular fingerprints or sequence data. HRGCN+~\cite{wu2021hyperbolic} is a notable advancement in this direction that combines graph representations with molecular descriptors using a hyperbolic relational GCN. By learning graph representations on Riemannian manifolds with differentiable exponential and logarithmic maps, HRGCN+ leverages the complementary nature of graph and descriptor-based representations. The research findings indicate that descriptors can significantly enhance the performance of graph-based methods on small datasets. Integrating different types of molecular representations in hybrid methods holds promise for addressing the limitations of individual techniques and improving overall prediction accuracy in MPP.
 
This conformational information significantly impacts geometric characteristics such as bond angles, dihedral angles, and bond lengths. Integrating conformational information into molecular representations allows for a more accurate depiction of a molecule shape and geometry, enhancing the precision and efficacy of MPP when combined with spatial data. Lu et al.~\cite{lu2019molecular} introduced a multilevel graph convolutional neural network that learns node representations while preserving both conformational and spatial information. Similarly, Liu et al.~\cite{liu2021spherical} proposed spherical message passing and SphereNet for 3D molecular learning. By leveraging relative 3D information and torsion computation, the model achieves generalized predictions of rotation and translation invariance. Likewise, Liu et al.~\cite{liu2021pre} also introduced a semi-supervised learning method that utilizes both 3D and 2D graphs of the same molecule for model training, teaching the model to derive 3D conformers from their 2D structures.

However, effectively handling conformational flexibility and capturing relevant conformers for property prediction pose computational challenges. To address this, researchers are increasingly incorporating curvature information to enhance the effectiveness of GNNs. Inspired by the Graphformer architecture~\cite{ying2021transformers}, Chen et al.~\cite{chen2023curvature} developed a curvature-based transformer model aimed at improving the capabilities of graph transformer neural network models. Their work demonstrates that discretized Ricci curvature preserves both structural and functional information along with local geometry within molecular graphs. The line graph transformer (LiGhT)~\cite{li2022kpgt} is an innovative application of transformer models tailored specifically for capturing the structural information of molecular graphs. This high-capacity model places particular emphasis on the significance of chemical bonds within molecules. These advancements signify a shift towards enhancing the structural understanding and predictive capabilities of GNN models in MPP.
\begin{table*}[ht!]
    \begin{center}
    \caption{An overview of the multi modality based methods for MPP}
    \label{Hybrid summary}
    \scalebox{0.70}{
    \begin{tabular}{p{1.0cm} p{3.0cm} p{1.8cm} p{3.2cm} p{2cm} p{2.5cm} p{3.0cm} c }\hline
    \label{tab:hybridleader}
    \textbf{Year} & \textbf{Dataset} & \textbf{Task}  & \textbf{Input Modalities} & \textbf{Method} &  \textbf{Evaluation} & \textbf{GitHub/ \newline Server} & \textbf{Reference}\\
    \hline
    \multirow{3}{*}{2019}
    & ESOL,Lipo,FreeSolv & Regression & Graph,Fingerprints & C-SGEL & Random & wxfsd/C-SGEN & Wang et al.~\cite{wang2019molecule}\\
   
   & IGC50 & Regression & SMILES,Descriptors,\newline Molecular Image & CNN,RNN,\newline FCNN,EA & 5-Fold CV & - & Karim et al.~\cite{karim2019toxicity}\\
    
    \hline
    \multirow{2}{*}{2020}
     & LogP,FDA,BBBP, \newline BACE,Tox21,ToxCast &  Classification & Graph,SMILES &  GNN,Bi-LSTM & Random,\newline Scaffold & - & Guo et al.~\cite{guo2020graseq}\\
    
   & hERG &  Classification & Descriptors,Fingerprints & DeepHIT & Random & - &  Ryu et al.~\cite{ryu2020deephit}\\
    \hline
    \multirow{4}{*}{2021}
    & ESOL,Lipo,FreeSolv,\newline HIV,BACE,BBBP,\newline Tox21,ToxCast,ClinTox,\newline SIDER & Regression,\newline Classification & Graph & GCN,GGNN,\newline DMPNN,\newline XGBoost & Stratified & chenxiaowei-vincent/XGraphBoost.git & Deng et al.~\cite{deng2021xgraphboost}\\
    
    
     & QM9,PC9 & Regression & Graph & MPNN ensemble & Random & - & Busk et al.\cite{busk2021calibrated}\\
    \hline
    \multirow{6}{*}{2022}
    & BACE,HIV,MUV,\newline Tox21, BBBP,Clintox,\newline SIDER & Regression,\newline Classification & Graph,MACCS,Pubchem,\newline Pharmacophores & FP-GNN & Random,\newline Scaffold & idrugLab/FP-GNN &  Cai et al.~\cite{cai2022fp}\\
    
   & BBBP & Classification & Descriptors,MACCS,\newline Molecular Image & DNN, 1D-CNN,\newline VGG-16   & Random  & - &  Kumar et al.~\cite{kumar2022deepred}\\
    
     & HIV,BACE,Lipo,\newline BBBP,ESOL,QM7,\newline FreeSolv & Classification,\newline Regression & SMILES,Descriptors & CNN,DNN,\newline Bayesian Optimization & Random & - & Chen and Tseng~\cite{chen2022general}\\
    
   & BBBP & Classification & Descriptors,Fingerprints,\newline  Molecular Graph,\newline SMILES & ResNet-50,\newline LSTM & Random & - & Tang et al.~\cite{tang2022merged}\\
    
    & BBBP & Classification & Molecular Graph,\newline Descriptors & RGCN & Stratified 10-Fold CV & - & Ding et al.~\cite{ding2022relational}\\

     
    \hline
    \multirow{6}{*}{2023}
    & Tox21 & Classification & Graph,SMILES  & GraphSAGE,\newline BiGRU & Random & jpliuhaha/jpliuhaha.git & Liu et al.~\cite{liu2023prediction}\\

    & ESOL,FreeSolv,\newline Lipophilicity, ClinTox,\newline BBBP,BACE & Classification,\newline Regression & Graph,Descriptors & GAT,DNN & 5-Fold CV & magdalenawi & Wiercioch and Kirchmair~\cite{wiercioch2023dnn}\\
    
   & hERG & Clasification & ECFP-2, PubChem, \newline AtomPairFingerprintCount, Molecule Graph & SGAT,DNN & 5-Fold CV & zhaoqi106/DMFGAM &  Wang et al.~\cite{wang2023investigating}\\
    
   & HIV,BACE,Lipo,\newline Tox21,ESOL,FreeSolv & Classification,\newline Regression & Descriptors,\newline Molecular Graph & DNN,GCN,GAT & Random & - & He et al.~\cite{he2023novel}\\
    
   & ESOL,Lipo,BACE & Classification,\newline Regression & ECFP,SMILES\newline Molecular Graph & Transformer Encoder, \newline GRU,GCN & Random & - & Lu et al. ~\cite{lu2023integrating}\\

     & HIV,BACE,Lipo,\newline BBBP,LogP,Tox21,\newline SIDER & Classification & SMILES,\newline Molecular Graph & MCNN,GIN,GRU & Random, Scaffold & - & Wu et al.~\cite{wu2023improved}\\

    & HIV,BACE,BBBP,\newline Clintox,QM9 & Classification & SMILES,ECFP,\newline Molecular Graph & MPNN,\newline MLP, Bi-LSTM & Random & - & Zheng et al.~\cite{zheng2023emppnet}\\
    \hline
    \multirow{1}{*}{2024}
    & Lipo,BACE,BBBP,\newline FreeSolv,Clintox,ESOL & Classification & ECFP,\newline Molecular Graph,\newline Molecular Image & FCNN,GCN,\newline EGNN, \newline VGG-16 & Random and Scaffold & - & Ma and Lie~\cite{ma2024deep}\\
    
    & Lipo,BACE,BBBP,\newline FreeSolv,Clintox,ESOL & Classification & Fingerprints,\newline Molecular Graph & FCNN,GNN & Scaffold & learningmatter- mit/geom & Ma et al.~\cite{zhang2024pre}\\
    
\hline
\end{tabular}}
 \end{center}
\end{table*}

%% file: nn_and_ls.tex
\section{Neural Networks and Learning Schemes}
\label{sec:nn_ls}
In this section, we explore the methodologies used in modeling MPP from two perspectives: 1) detailing the modules used as foundational elements for shaping MPP architectures, and 2) outlining the learning schemes used to train models for better generalization.
\subsection{Key Considerations for Implementing NN Models for MPP}
Building a neural network architecture requires careful selection of components, fine-tuning their parameters, and structuring them into a cohesive network layout. Deep learning frameworks offer a range of modules, from fundamental dense layers to more complex architectures like GNNs, Transformers, and RNNs. Although there isn't a standardized way for designing pipelines, the selection of modules can be informed by the characteristics of the input data. For instance, GNNs are commonly used for graph-structured data to capture relational information. Transformers leverage self-attention mechanisms and excel in sequential data tasks while RNNs are suitable for processing data with temporal and sequential dependencies. Building upon the insights from Sultan et al.~\cite{sultan2024transformers}, this section illustrates the fundamental concepts of these architectures and delve into the specific decisions required for constructing and training these models for MPP. The detailing about the architecture is illustrated in Appendix A.
\begin{figure}[ht!] 
\centering
\includegraphics[scale = 0.45]{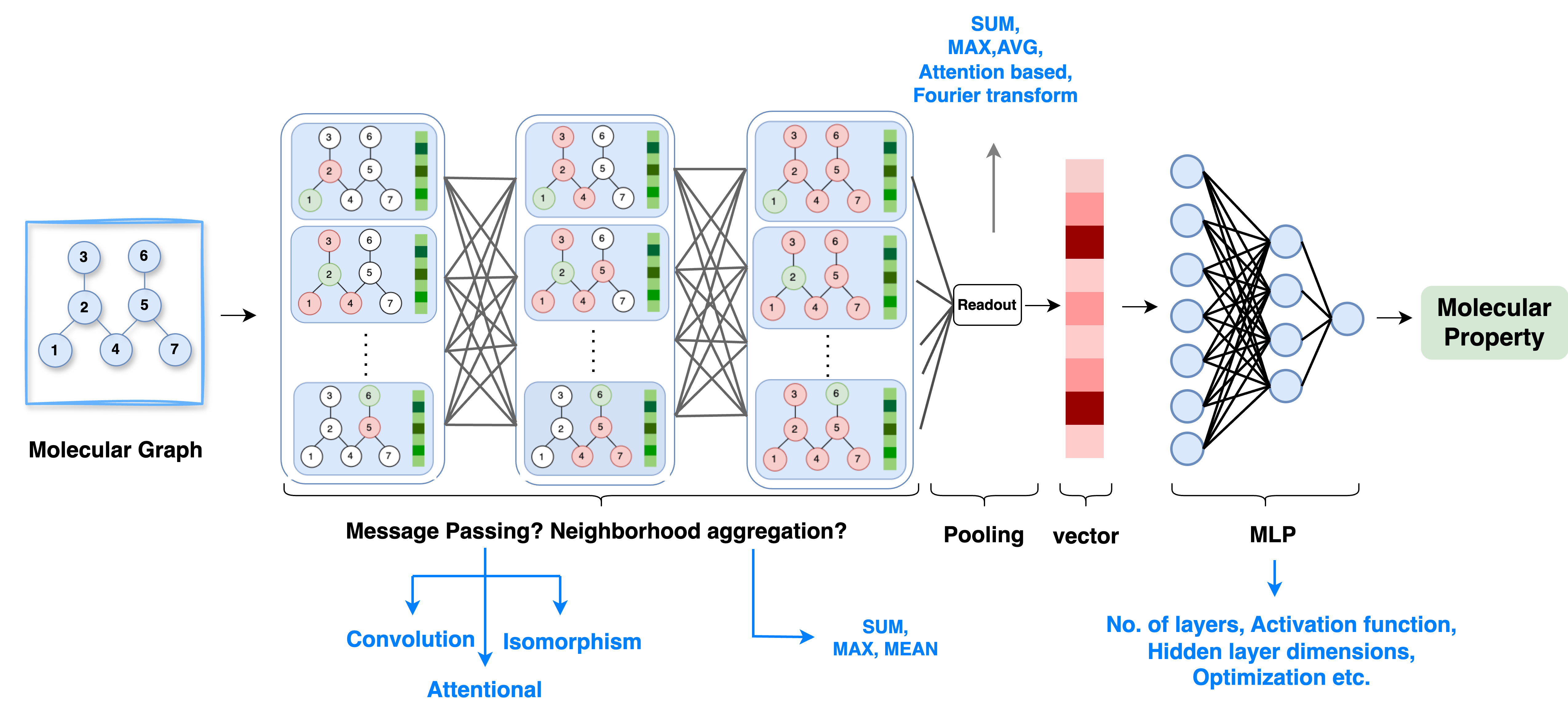}
\caption{Various decision points to be considered during the construction of GNN in MPP}
\label{fig:DD GNN}
\end{figure}
\subsubsection{GNN}
The construction of a GNN involves making decisions regarding the message passing mechanism, update function, readout operation, and architectural parameters to optimize the model's performance and effectiveness in capturing relevant molecular features as illustrated in Figure~\ref{fig:DD GNN}. We identified the decision points based on GNN architecture with possible options that a user can take while constructing and training a GNN.
\begin{enumerate}
    \item The choice of message passing mechanism depends on the characteristics of the molecular data and the specific task requirements. For example, attention mechanisms are effective for capturing long-range dependencies, while convolutional operations are suitable for capturing local structural patterns. The three different message passing mechanisms - Convolutional, Isomorphism, and Attentional can be explored based on the requirement. 
    \item After the message passing phase, a readout operation can be done using different methods. Some common readout operations include: 
    \begin{itemize}
    \item \textbf{Sum:} The node embeddings are simply summed to obtain the graph representation.
    \begin{equation}
        h_{graph} = \sum \limits_{v\in V}^{} h_v
    \end{equation}
    \item \textbf{Mean:} Similar to the sum readout, but instead of summing, the mean of all node embeddings is computed. 
    \begin{equation}
        h_{graph} = \frac{1}{|v|} \sum \limits_{v\in V}^{} h_v
    \end{equation}
    \item \textbf{Max:} The maximum value of each dimension across all node embeddings is taken to obtain the graph representation.
    \begin{equation}
        h_{graph} = \max \limits_{v\in V}^{} h_v
    \end{equation}
    \item \textbf{Attention-based readout:} An attention mechanism can be applied to assign importance weights to node embeddings before aggregating them.
     \begin{equation}
        \phi_{v} = \sigma \left( MLP \left(h_v\right)\right)
    \end{equation}
    \end{itemize}
    The selection of the readout operation influences the final graph-level representation and consequently the performance of the model in downstream tasks. 
    \item The architecture of GNN may includes multiple layers of message passing units. The decision regarding the number of layers, activation functions, hidden layer dimensions, and optimization methods significantly impacts the capacity of the model and learning capabilities. Experimentation and tuning to determine the optimal architecture for the given MPP task is also essential.
\end{enumerate}
\subsubsection{Recurrent Neural Networks}
To illustrates the decision-making process involved in designing a RNN for MPP, we identified the following decision pointers.
\begin{figure}[ht!] 
\centering
\includegraphics[scale = 0.65]{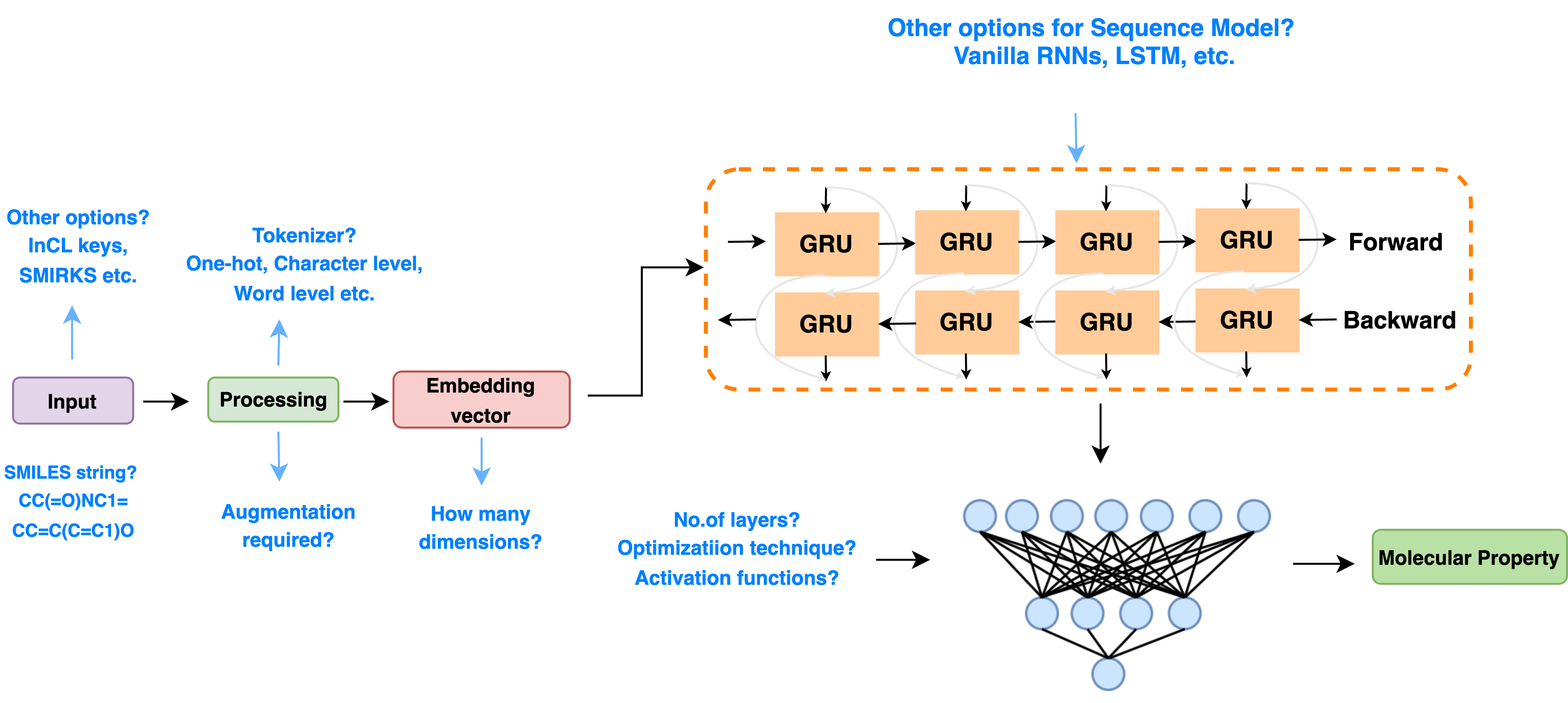}
\caption{Various decision points to be considered during the construction of sequence model in MPP}
\label{fig:DD RNN}
\end{figure}
\begin{enumerate}
    \item The input can be SMILES, SELFIES, etc. which need to be preprocessed before feeding into the RNN.
    \item Preprocessing the input data, which includes tokenization and embedding. Tokenization methods such as one-hot encoding, character-level encoding, or word-level encoding are chosen based on the requirement. Additionally, embedding techniques and augmentation strategies (if required) are applied to transform the tokenized data into suitable input vectors. Augmentation methods such as random insertion or deletion of atoms or bonds, flipping or rotating molecules, introducing noise or perturbations, and generating tautomers can be introduced in a controlled way preserving the molecular validity of the augmented samples.
    \item The RNN architecture is a critical decision point with options like vanilla RNNs, LSTM, or GRU being considered. Each architecture has its strengths and weaknesses, and the choice depends on factors such as the complexity of the MPP task and computational resources.
    \item Decision on the architectural details is important which include the number of layers, selection of activation functions, and optimization techniques. These choices collectively define the model's complexity, learning capacity, and training dynamics, and are crucial for achieving optimal performance in MPP tasks.
\end{enumerate}
\subsubsection{Transformers}
\begin{figure}[ht!] 
\centering
 \includegraphics[scale = 0.65]{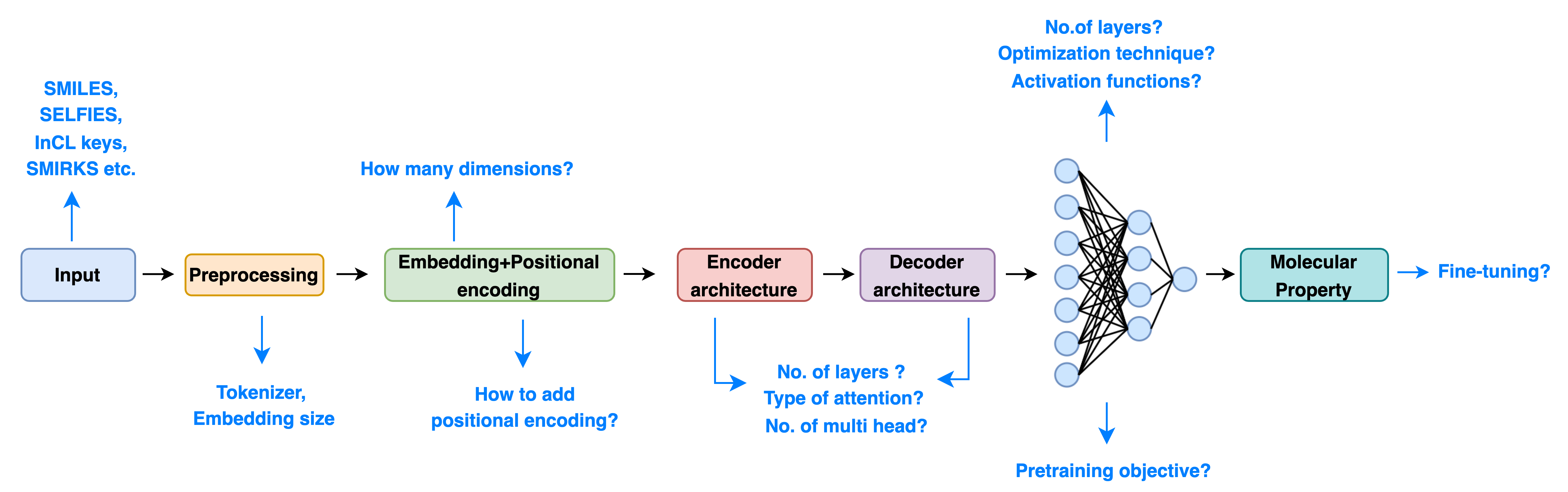}
  \caption{Various decision points to be considered during the construction of transformer in MPP}
  \label{fig:DD transformer}
\end{figure}

 Figure \ref{fig:DD transformer} outlines various decision points involved in designing a transformer model for MPP. The illustration of the decision points involved in transformer is given below.\\

\begin{enumerate}
    \item Decision on the input representation such as SMILES, SELFIES, InChI keys, SMIRKS, etc. is crucial. It involves selecting the most suitable molecular representation for the given MPP task.
    \item Determining the preprocessing steps required for the input data, such as tokenization and embedding. This decision includes choices like tokenizing the input, defining the length of each embedding.
    \item Deciding on the method for creating embeddings and adding positional encoding to the input data. This step ensures that the model can effectively capture the spatial relationships between different parts of the input molecules.
    \item Choosing the architecture of the encoder, including the number of layers, type of attention mechanism, and number of multi-head attention heads. These decisions shape how information is processed and propagated through the model.
    \item Similar to the encoder, deciding on the architecture of the decoder, including the number of layers and type of attention mechanism. The decoder is responsible for generating the output based on the encoded input information.
    \item Determining the configuration of the feedforward layers within the transformer model, including the number of layers, optimization technique, and activation functions. 
    \item Deciding whether fine-tuning is necessary for the output layer of the transformer model. Fine-tuning allows the model to adapt its parameters to better fit the specific MPP task at hand.
\end{enumerate}
\subsection{Learning Schemes}
Various learning schemes, such as transfer learning, ensemble methods, and semi-supervised learning, enhance model performance. These techniques apply independently of the specific neural network components used in the pipeline. Transfer learning pre-trains a neural network on a large, similar dataset before fine-tuning it on a smaller, domain-specific dataset. Ensemble methods combine multiple models to improve predictive performance, regardless of their architectures. The subsequent section details these learning schemes.
\subsubsection{Transfer Learning}
Transfer learning is a powerful technique that repurposes a model trained on one task for a related task, leveraging knowledge gained from solving one problem to apply it to a different but related problem. This approach is particularly valuable when labeled data for the target task is scarce, as it allows the model to benefit from the larger dataset available for the source task.\\ 
In the domain of drug discovery, transfer learning, combined with DL has demonstrated significant success~\cite{cai2020transfer}. For example, Shen and Nicolaou~\cite{shen2019molecular} used transfer learning to predict drug candidate permeability, achieving notable advancements. Hu et al.~\cite{hu2020gpt} pre-trained a robust GNN model through self-supervision on unlabeled data, then used this pre-trained model for downstream tasks. GNNs excel in transfer learning for MPP due to their ability to capture complex relational information in graph-structured data.\\
Pretraining a GNN at both node and graph levels yields valuable local and global representations, consistently improving performance across various molecular property datasets~\cite{hu2019strategies}. Li and Rangarajan~\cite{li2022conceptual} highlighted that the success of transferring knowledge from a source model to a target model depends on the similarity between tasks. Greater feature overlap between tasks enhances transfer learning success. Buterez et al.~\cite{buterez2024transfer} trained a GNN on low-fidelity data to learn molecular representations and then fine-tuned it to improve predictions on high-fidelity data with limited labeled samples. Li et al.~\cite{li2022improving} introduced MoTSE, pre-training GNN models on task-specific datasets. MoTSE uses attribution and molecular representation similarity analysis to project tasks into a unified latent space, estimating task similarity based on vector distances in this space for quantitative comparison. 
\subsubsection{Ensemble Models}
Ensemble methods like bagging, boosting, or stacking are widely used to aggregate diverse model predictions, improving generalization and predictive accuracy in MPP~\cite{liu2021prediction}. Hu et al.~\cite{hu2021development} developed a Hildebrand solubility prediction model using ensemble of random forest, gradient boosting, and extreme gradient boosting. It can also utilize DL algorithms such as GNNs, CNNs, RNNs, depending on data characteristics and specific tasks~\cite{kosasih2021graph, busk2021calibrated}. Ensemble learning has shown effective performance in predicting carcinogenicity and identifying structural features linked to carcinogenic effects~\cite{zhang2017carcinopred}. By harnessing the collective intelligence of multiple models, ensemble methods mitigate individual biases and uncertainties, resulting in more robust predictions for molecular properties.
\subsubsection{Contrastive Learning}
Contrastive learning trains models to distinguish molecular structures by maximizing similarity between representations of similar molecules and minimizing it for dissimilar ones. This self-supervised technique uses positive pairs (similar molecules) and negative pairs (randomly selected dissimilar molecules) to learn from unlabeled data by penalizing errors, forcing the model to differentiate between them during training. Graph contrastive learning (GCL) methods have shown promise in scenarios with limited labeled data, employing tailored data augmentation strategies for graphs~\cite{guan2023t, zheng2023casangcl, pinheiro2022smiclr}. Liu et al.~\cite{liu2022attention} introduced attention-wise graph masking to create challenging positive samples. However, conventional augmentation techniques like random perturbation may unintentionally alter molecular properties, leading to representation conflicts and training imbalances. To address these issues, He et al.~\cite{he2023novel} proposed a two-stage framework. In the first stage, they use contrastive learning to pre-train encoders for descriptors and graph representations, enhancing representational consistency. The second stage employs supervised learning with a multi-branch predictor architecture to integrate target attribute labels, improving decision fusion and addressing training imbalances. The role of 3D structures is crucial for capturing detailed spatial relationships and atomic configurations within molecules. Moon et al.~\cite{moon20233d} and Kuang et al.~\cite{kuang20233d} emphasized the importance of 3D information in graph contrastive frameworks. Moon et al. utilized a conformer pool to maintain molecular integrity during contrastive learning, while Kuang et al. developed an encoder for extracting 3D features by decomposing molecules into geometric graphs. Recently, Li et al.~\cite{li2022geomgcl} introduced GeomGCL, a novel graph contrastive learning method (GeomMPNN) that leverages both 2D and 3D views of molecules which effectively integrate geometric information from multiple perspectives. This approach enhances the representation learning capabilities of graph-based models, promising more accurate and robust predictive models in computational chemistry.
\begin{figure}[ht!] 
\centering
\includegraphics[scale = 0.30]{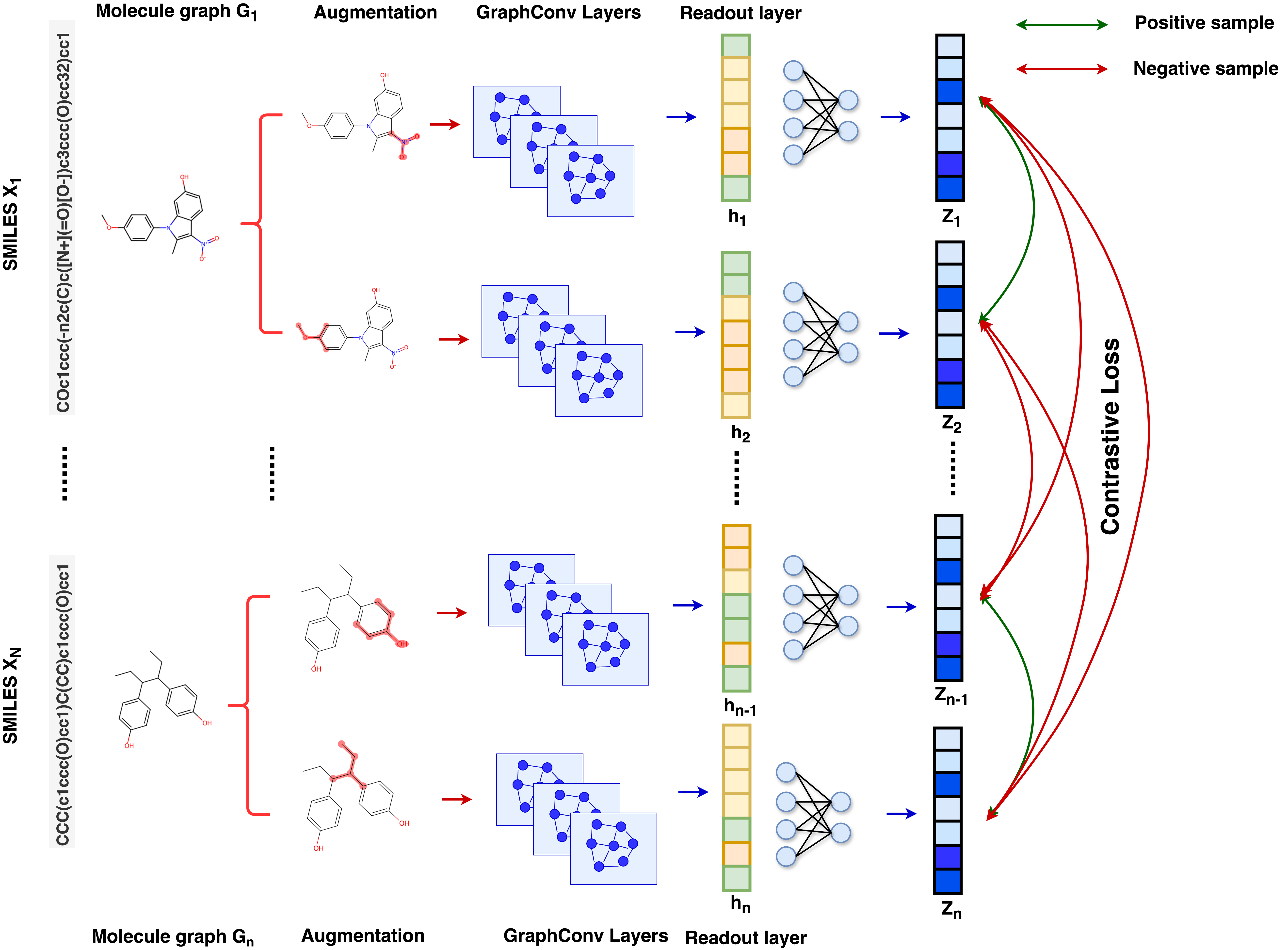}
\caption{The illustration of contrastive learning using graph convolution layers}
\label{multitasking}
\end{figure}
\subsubsection{Few-shot learning}
Few-shot learning in machine learning trains models with a limited number of labeled examples. Unlike traditional supervised learning, which requires abundant labeled data, few-shot learning generalizes from a small number of examples, making it suitable where acquiring labeled data is costly. 
Torres et al.~\cite{torres2023few, torres2023few1} introduces a meta-learning framework that iteratively updates model parameters across few-shot tasks to predict new molecular properties using limited data. However, challenges persist in methods based on GNNs because finding molecules with desired properties remains difficult. Hierarchically Structured Learning on Relation Graphs (HSL-RG)~\cite{ju2023few} focuses on capturing molecular structural semantics using graph kernels and self-supervised learning, adapting meta-learning for customized predictions with limited data. In QSPR studies, specific molecular properties correlate with distinct substructures, varying across different tasks like in Tox21 dataset. Wang et al.~\cite{wang2021property} address these challenges with an adaptive relation graph learning module refining molecular embeddings via few-shot learning tailored to target properties. Future research will explore integrating GNNs into few-shot learning, leveraging molecular graph representations for enhanced predictive performance.
\subsubsection{Multi-Task learning}
Multi-Task Learning (MTL) trains a model to handle multiple related tasks simultaneously, leveraging shared knowledge across tasks to boost overall performance. 
In MPP, MTL involves training a single model to predict multiple molecular properties concurrently. This approach aims to streamline model development by reducing the need for separate models per task, optimizing computational resources. However, selecting the right molecular descriptors and fingerprints remains critical for effective model performance, posing significant challenges in model development. To address these challenges, Lim and Lee~\cite{lim2021predicting} explored transformer-based models with self-attention mechanisms within an MTL framework, demonstrating the effectiveness of self-attention in improving molecular representation learning across diverse chemical tasks. Studies~\cite{liu2022structured} have also investigated MTL using structured relation graphs between tasks, highlighting its potential to leverage data across tasks and handle data scarcity. 
MTL-BERT~\cite{zhang2022pushing} integrates extensive pre-training, MTL, and SMILES enumeration to address data scarcity effectively. Overall, the ability of MTL to exploit shared information, regularize learning, improve data efficiency, facilitate knowledge transfer, and capture task correlations. 

%% file: resource_availability.tex
\section{Resource Availability}\label{sec:resource_availability}
\subsection{Datasets Overview}
\begin{figure}[ht!] 
\centering
 \includegraphics[scale = 1.2]{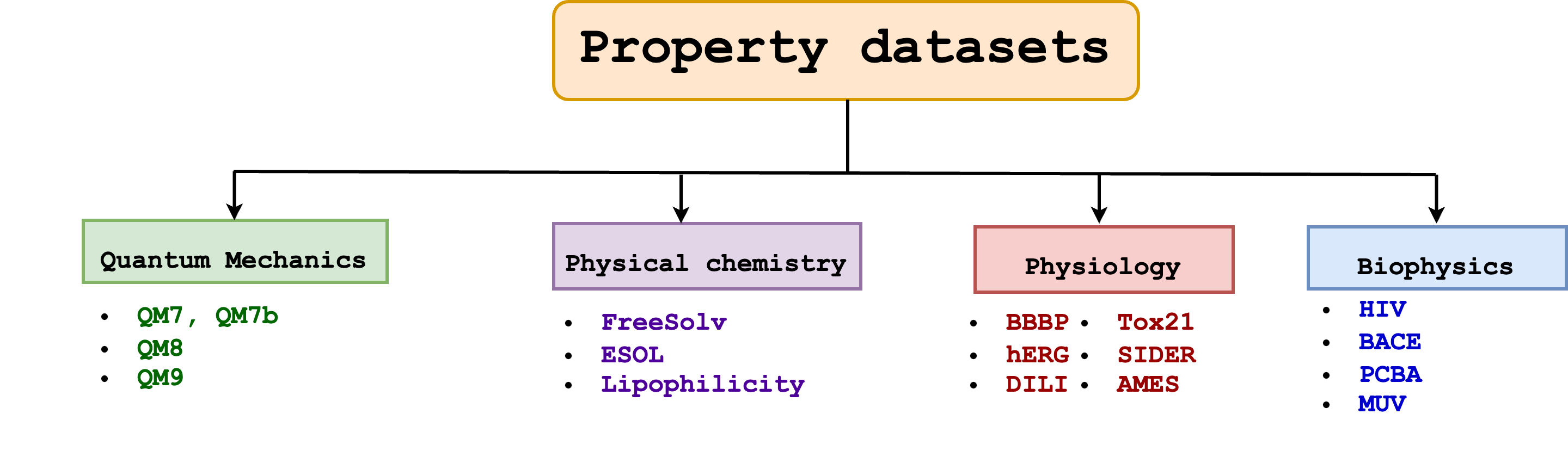}
  \caption{Categorization of MPP datasets}
  \label{fig:dataset}
\end{figure}

We present an overview of the categories of MPP datasets in Figure~\ref{fig:dataset}. Each category consists of multiple datasets. 
Many of these datasets are integral components of the MoleculeNet benchmark, a valuable resource for evaluating ML methods in molecular ML and cheminformatics ~\cite{wu2018moleculenet}. MoleculeNet serves as a pivotal tool for research, development, and comparison of diverse algorithms designed for property prediction tasks. It offers a standardized collection of datasets along with consistent evaluation processes and tools. The datasets within MoleculeNet encompass over 700,000 compounds, covering properties across four main categories: quantum mechanics, biophysics, physical chemistry, and physiology. Quantum mechanics datasets focus on the electronic properties of compounds and include QM7, QM7b, QM8, and QM9 datasets. Physical chemistry datasets are centered around thermodynamic concepts, featuring data on hydration free energy, solubility, and octanol/water distribution coefficients. Biophysics datasets provide information on biological properties such as binding affinities (e.g., BACE), interactions, and efficacies. Physiology datasets contain insights into drug side-effects, drug-related toxicological effects, and more. Although datasets like Ames and hERG are not part of MoleculeNet, they still represent valuable contributions from researchers focused on mutagenicity and cardiotoxicity prediction.

Table~\ref{tab:datasets_summary} offers a detailed summary of the datasets used in the reviewed article, including their statistics, summary, and accessibility links.Notably, we have included the link to Hansen's Ames mutagenicity dataset~\cite{hansen2009benchmark}, widely recognized as a benchmark in the field. While several datasets curated by various researchers exist for predicting cardiotoxicity, we have chosen to highlight Karim's dataset~\cite{karim2019toxicity, karim2021cardiotox} due to its relevance to recent studies.
\subsection{Computational tools and servers}
 To provide valuable insights into the computational resources available for similar investigations, we present a comprehensive list of tools and servers in Table \ref{tab:server_summary}. This table outlines various aspects such as the number of descriptors and fingerprints, the dimensionality and type of descriptors provided, accessibility via access links, and availability of graphical interfaces.

 \begin{table}[ht!]
    \centering
    \caption{Summary of Datasets in MPP}
    \scalebox{0.58}{
    \begin{tabular}[t]{c p{2.0cm} p{11.0cm} c c c}\hline
    \label{tab:datasets_summary}
    \textbf{Property} & \textbf{Dataset/Link} & \textbf{Description} & \textbf{Tasks} & \textbf{Compounds} & \textbf{Reference}\\
    \hline
    \\
    Solubility & ESOL~\textsuperscript{*} & contains chemical structures along with their corresponding experimentally determined solubility values in water & 1 & 1128  &~\cite{chen2022general}\\
    Solv. Energy & FreeSolv~\textsuperscript{*} & provide chemical structures along with experimentally determined solvation free energy values & 1 & 643 &~\cite{li2022novel}\\
    Hydrophobicity, Hydrophilicity  & Lipophilicity~\textsuperscript{*} & depicts the tendency of a molecule to dissolve in lipids or non-polar solvents  & 1 & 4200 &~\cite{tian2023predicting}\\
    Affinity & BACE~\textsuperscript{*} & consists of molecules that are tested for their ability to inhibit the BACE enzyme, with associated experimental measurements of their inhibitory activity & 1 & 1522 & \cite{yang2022ensemble}\\
    Permeabiltiy & BBBP\textsuperscript{*} & contains molecular structures of compounds along with their experimentally measured blood-brain barrier permeability values & 1 & 2053 &~\cite{tang2022merged}\\
    Toxicity & Tox21~\textsuperscript{*} & contains information on the biological activity of thousands of compounds across a panel of assays covering a range of biological processes and endpoints & 12  & 8014 &~\cite{liu2021pre}\\
    Toxicity & Toxcast~\textsuperscript{*} & includes results from various assays covering endpoints related to cytotoxicity, genotoxicity etc. and the biological activity of chemicals across them & 617 & 8615 &~\cite{tian2023predicting}\\
    Toxicity & ClinTox~\textsuperscript{*} & consists of molecular structures along with binary labels indicating whether each molecule is associated with toxicity or not & 2 & 1491 &~\cite{wiercioch2023dnn}\\
    Side-effects & SIDER~\textsuperscript{*} & provides comprehensive coverage of adverse drug reactions (ADRs) associated with a wide range of drugs & 27 & 1427 &~\cite{tian2023predicting}\\
    Quantum mechanics & QM7~\textsuperscript{*} & includes atomization energies and energies of the highest occupied molecular orbital (HOMO) for organic molecules & 1 & 7165 &~\cite{shindo2019gated}\\
    Bioactivity & MUV~\textsuperscript{*} &  contains a set of molecules labeled as active (binders) or inactive (non-binders) with respect to the 17 different biological targets & 17 & 93127 &~\cite{li2017learning}\\
    Efficacy & HIV~\textsuperscript{*}& provide results from screening experiments aimed at identifying compounds with potential anti-HIV activity & 1 & 41913 &~\cite{li2017learning} \\
  
    Cardiotoxicity & hERG blocking~\textsuperscript{**} & contains information about the inhibition activity of compounds against the hERG potassium ion channel & 1 & 12620 &~\cite{wang2023investigating} \\
 
    Mutagenicity & Ames data~\textsuperscript{***} & contains structural information and experimental results for a large number of chemical compounds tested for their mutagenic activity & 1 & 5395 &~\cite{chu2021machine} \\
  
     Mutagenicity & Hansen data~\textsuperscript{****} & contains SMILES and experimental results for a large number of chemical compounds tested for their mutagenic activity & 1 & 6277 &~\cite{hansen2009benchmark} \\
    \hline
\end{tabular}}
\tiny
\begin{tablenotes}
    \setlength{\parskip}{0pt}%
    \item[*]\url{https://moleculenet.org/datasets-1}
    \item[**]\url{https://github.com/zhaoqi106/DMFGAM}
    \item[***]\url{https://pubs.acs.org/doi/10.1021/ci300400a}
    \item[****]\url{https://pubs.acs.org/doi/abs/10.1021/ci900161g}   
\end{tablenotes}
\end{table}

\begin{table*}[ht!]
    \centering
    \caption{Various descriptor calculation packages/servers and their comparison}
    \scalebox{0.55}{
    \begin{tabular}{c c c c c c}\hline
    \label{tab:server_summary}
    \textbf{Package/Server} & \textbf{Descriptors} & \textbf{Citation count} &  \textbf{Type} & \textbf{GUI} & \textbf{Access link}\\
    \hline
    Mordred & 1826 Descriptors & 643 & 2D and 3D & - & \href{https://pypi.org/project/mordred}{https://pypi.org/project/mordred} \\
    Chemdes & 3679 descriptors, 59 fingerprints &  264 & 1D, 2D and 3D & \checkmark & \href{www.scbdd.com/chemdes}{www.scbdd.com/chemdes} \\
    PaDELpy & 1875 descriptors, 12 fingerprints & 2258 &  1444 1D, 2D, and 431 3D & \checkmark  & \href{http://www.yapcwsoft.com/dd/padeldescriptor}{http://www.yapcwsoft.com/dd/padeldescriptor} \\
    CDK\_pywrapper & - & - & 1D, 2D, 3D descriptors and fingerprints  & - & \href{https://pypi.org/project/CDK-pywrapper/}{https://pypi.org/project/CDK-pywrapper/} \\
    pybel & - & 408 & 1D, 2D descriptors & - & \href{https://pypi.org/project/pybel/}{https://pypi.org/project/pybel/} \\
    
    PyBioMed & 775 descriptors, 19 fingerprints & 112 & 1D, 2D, 3D descriptors & \checkmark &  \href{http://projects.scbdd.com/pybiomed.html}{http://projects.scbdd.com/pybiomed.htm}\\
    Rcpi & >300 molecular descriptors and 10 fingerprints & 130 & 1D,2D Descriptors & - & \href{http://bioconductor.org/packages/release/bioc/html/Rcpi.html} {http://bioconductor.org/packages/release/bioc/html/Rcpi.html}\\
    Biotriangle & 540 descriptors and 7 fingerprints & 47 & 1D, 2D descriptors & \checkmark & \href{http://biotriangle.scbdd.com}{http://biotriangle.scbdd.com}\\

    \hline
\end{tabular}}
\end{table*}

%% file: comparative_analysis.tex
\section{Comparative Analysis of State-of-the-Art Methods and Evaluation Performance}
\label{sec:CA and EP}
To provide valuable context within the existing literature, we present a comprehensive overview of the top state-of-the-art (SOTA) methods for common MPP datasets. This includes detailed information on their evaluation performance and splitting criteria, as outlined in Tables~\ref{tab:MPP_Class_T5} and ~\ref{tab:MPP_Regr_T5}. By offering this comparative analysis, our objective is to facilitate model selection, promote transparency, and provide insights into the advantages and limitations of different approaches. Among the top methods presented in Table~\ref{tab:MPP_Class_T5} and ~\ref{tab:MPP_Regr_T5}, it becomes evident that molecular graph-based and SMILES-based methods consistently demonstrate superior performance compared to other approaches across multiple property datasets. This dominance underscores the efficacy of using molecular graph representations for property prediction tasks. Graph-based methods excel in capturing the intricate structural relationships present in molecules. The comprehensive performance exhibited by molecular graph-based models across various datasets highlights their robustness and versatility in handling diverse molecular structures and properties. For instance, Attentive-FP~\cite{xiong2019pushing}, a graph based approach has shown significant performance over datasets such as ESOL, FreeSolv, Lipophilicity, BBBP, HIV, and Clintox. For datasets like ESOL, FreeSolv, and Lipophilicity, graph-based methods have demonstrated dominance in performance over other modalities in general. Graph-based approaches, which capture geometric information in addition to structural and topological features, have shown significant performance improvements, as evidenced by one of the studies conducted by Chen et al.~\cite{chen2023curvature}. On the contrary, for classification datasets such as  BBBP, HIV, Tox21, Clintox, SIDER, and Toxcast, SMILES-based methods have also exhibited significant performance gains. Additionally, for datasets like AMES and hERG channel blocker, multi-modality methods have shown promising performance. To ensure fairness and consistency in comparing methods for hERG blocking prediction, we report the evaluation performance exclusively for Karim's test dataset I in Table~\ref{tab:MPP_Class_T5}. The number of studies utilizing multiple modalities is relatively lower compared to those focusing on single modality methods. This highlights an area for potential future research exploration, as multi-modality methods hold promise for further enhancing the predictive capabilities of prediction models.

\begin{table}[h!]
  \begin{center}
    \caption{Performance evaluation of selected models for the standard molecular property classification datasets}
    \label{tab:MPP_Class_T5}
    \scalebox{0.68}{
    \begin{tabular}{c c c c c c c}
    \hline
      \textbf{Dataset} & \textbf{Model/Reference} & \textbf{Year} & \textbf{Input} & \textbf{Information Type} & \textbf{Splitting} & \textbf{AUC} \\
      
      \hline
      \multirow{5}{*}{Ames} &  Shinada et al.~\cite{shinada2022optimizing} & 2022 & Descriptors \& Fingerprints & Topological \& Structural & 5-CV & \textbf{0.93}\\
      & Winter et al.~\cite{winter2019learning} & 2019 & SMILES & Structural & Random & 0.89\\
      & Zhang at el.~\cite{zhang2017novel} & 2017 & Descriptors \& Fingerprints & Topological \& Structural & 5-CV & 0.89\\
      & karim et al.~\cite{karim2019efficient} & 2019 & Descriptors & Structural & Random & 0.879\\
      \hline
      \multirow{4}{*}{hERG} &  CardioTox net~\cite{karim2021cardiotox} & 2021
      & Graph features, SMILES, Descriptors \& Fingerprints & Topological \& Structural & 10-CV & \textbf{0.81}\\
      & Shan et al.~\cite{shan2022predicting} & 2022 & Fingerprints \& Descriptors & Structural & Random \& Scaffold & 0.80\\
      & DMFGAM~\cite{wang2023investigating}
      & 2023 & Molecular graph \& Fingerprints & Topological \& Structural & 5-CV & 0.795\\
      \hline      
      \multirow{5}{*}{BBBP} & SA-MTL~\cite{lim2021predicting} & 2021 & SMILES & Structural & Scaffold & \textbf{0.954} \\

      & GraSeq~\cite{guo2020graseq} & 2020 & Molecular Graph \& Molecular Sequence & Topological \& Structural & Scaffold & 0.9426 \\
      
      & FP-GNN~\cite{cai2022fp} & 2022 & Molecular Graph \& Fingerprints & Sub-Structural \& Topological & Random & 0.935 \\
      
      & HRGCN+~\cite{wu2021hyperbolic} & 2021 & Molecular Graph \& Decriptors & Structural \& Geometrical & 50 Random splits & 0.926 \\
     
    
      \hline
      \multirow{5}{*}{HIV} & Li et al.~\cite{li2022novel} & 2022 & Multiple SMILES augmentation & Structural & 5-CV & \textbf{0.9767} \\

      & AttentiveFP~\cite{xiong2019pushing} & 2019 & Molecular Graph &  Structural & Scaffold & 0.832 \\
      
      & Transformer-CNN~\cite{karpov2020transformer} & 2020 & SMILES augmentation & Structural & 5-CV & 0.83 \\
      
      & SA-MTL~\cite{lim2021predicting} & 2021 & SMILES & Structural & Scaffold & 0.826 \\


      \hline
      \multirow{5}{*}{BACE} & CD-MVGNN~\cite{ma2022cross} & 2022 & Molecular Graph & Topological & Scaffold & 0.892 \\
      
      & Chen et al.~\cite{chen2023curvature} & 2023 & Molecular Graph & Structural\& Geometrical & Random & \textbf{0.889} \\

      & CLM~\cite{markert2020chemical} & 2020 & SMILES & Structural & Scaffold & 0.861 \\

      & FP-GNN~\cite{cai2022fp} & 2022 & Molecular Graph \& Fingerprints & Sub-Structural \& Topological & Scaffold & 0.86 \\



      \hline
      \multirow{5}{*}{Tox21} & Mol-BERT~\cite{li2021mol} & 2021 & SMILES & Structural & Scaffold & 0.923 \\

      & SA-MTL~\cite{lim2021predicting} & 2021 & SMILES & Structural & Random & 0.9 \\

      & CLM~\cite{markert2020chemical} & 2020 & SMILES & Structural & Random & 0.858 \\

      & HRGCN+~\cite{wu2021hyperbolic} & 2021 & Molecular Graph \& Decriptors & Structural \& Geometrical & 50 Random splits & 0.848 \\


      \hline
      \multirow{5}{*}{Clintox}   & SA-MTL~\cite{lim2021predicting} & 2021 & SMILES & Structural & Scaffold & \textbf{0.99}\\
      
      & CD-MVGNN~\cite{ma2022cross} & 2022 & Molecular Graph & Topological & Scaffold & 0.954 \\
        
      & Chen et al.~\cite{chen2023curvature} & 2023 & Molecular Graph & Structural\& Geometrical & Random & 0.941  \\

      &  AttentiveFP~\cite{xiong2019pushing} & 2019 & Molecular Graph &  Structural & Random & 0.94 \\

      \hline
      \multirow{5}{*}{SIDER} &  Mol-BERT~\cite{li2021mol} & 2021 & SMILES & Structural & Scaffold & \textbf{0.695} \\

      & FP-GNN~\cite{cai2022fp} & 2022 & Molecular Graph \& Fingerprints & Sub-Structural \& Topological & Scaffold & 0.661\\

      & CLM~\cite{markert2020chemical} & 2020 & SMILES & Structural & Random & 0.658\\

      & HRGCN+~\cite{wu2021hyperbolic} & 2021 & Molecular Graph \& Descriptors & Structural \& Geometrical & 50 Random splits & 0.641\\
      
      
      \hline
      \multirow{5}{*}{Toxcast} &  Transformer-CNN~\cite{karpov2020transformer} & 2020 & SMILES augmentation & Structural & 5-CV & \textbf{0.82}\\
      & XGraphBoost~\cite{deng2021xgraphboost} & 2021 & Molecular Graph & Topological & Stratified random split &  0.797 \\
      & TrimNet~\cite{li2021trimnet} & 2021 & Molecular Graph & Topological & Random &  0.777\\
      & GraSeq~\cite{guo2020graseq} & 2020 & Molecular Graph \& Molecular Sequence & Topological \& Structural & Random & 0.733\\
      \hline
      
      \end{tabular}}
  \end{center}
\end{table}
\begin{table}[h!]
  \begin{center}
    \caption{Performance evaluation of selected models for the standard molecular property regression datasets}
    \label{tab:MPP_Regr_T5}
    \scalebox{0.77}{
    \begin{tabular}{c c c c c c c}
    \hline
      \textbf{Dataset} & \textbf{Model/Reference} & \textbf{Year} & \textbf{Input} & \textbf{Information Type} & \textbf{Splitting} & \textbf{RMSE} \\

      \hline
      \multirow{5}{*}{ESOL} & Chen et al.~\cite{chen2023curvature} & 2023 & Molecular Graph & Structural\& Geometrical & Random & \textbf{0.493}  \\ 
      & Attentive-FP~\cite{xiong2019pushing} & 2019 & Molecular Graph & Structural & Random & 0.509  \\
      & FP- 
      BERT~\cite{wen2022fingerprints} & 2022 & SMILES \& ECFP-2 & Sub-structural & Random & 0.552 \\
      & HRGCN+~\cite{wu2021hyperbolic} & 2021 & Molecular Graph \& Descriptors & Structural \& Geometrical & Random  & 0.563 \\
      
      \hline
      \multirow{5}{*}{FreeSolv} & Attentive-FP~\cite{xiong2019pushing} & 2019 & Molecular Graph & Structural & Random & \textbf{0.736}\\ 
      & CGEN+FP~\cite{wang2019molecule} & 2019 & Molecular Graph \& Fingerprints & Structural \& Topological & Random & 0.78 \\
      
      & FP-GNN~\cite{cai2022fp} & 2022 & Molecular Graph \& Fingerprints & Sub-Structural \& Topological & Random & 0.905 \\ 
      
      & Transformer-CNN~\cite{karpov2020transformer} & 2020 & SMILES augmentation & Structural & 5-CV & 0.91 \\
      

      \hline
      \multirow{5}{*}{Lipophilicity} & IFGN~\cite{tian2023predicting} & 2023 & Molecular Graph & Topological & Scaffold & \textbf{0.574} \\

      & Attentive-FP~\cite{xiong2019pushing} & 2019 & Molecular Graph & Structural & Random & 0.578 \\ 
      & Maxsmi~\cite{kimber2021maxsmi} & 2021 & Augmented SMILES &  Structural & Train-Test & 0.592  \\

      & HRGCN+~\cite{wu2021hyperbolic} & 2021 & Molecular Graph \& Decriptors & Structural \& Geometrical & 50 Random splits & 0.603 \\ 
      
\hline
\end{tabular}}
  \end{center}
\end{table}

%% file: discussion.tex
\section{Discussion}\label{sec:discussion}
Despite significant advancements in computational techniques and the increased accessibility of extensive molecular datasets, several challenges persist in the field of MPP. These challenges center on exploring generalizability and transferability, navigating data quality and representation, enhancing interpretability, and managing the high dimensionality associated with integrating multimodal data. Given the persistent challenges in MPP, there are numerous opportunities to advance the field. These include  multitask learning to enhance model versatility across diverse tasks, uncertainty quantification techniques to assess prediction reliability, dimensionality reduction methods for optimizing computational efficiency, contrastive learning to improve feature representations, and explainable AI for transparent and interpretable model insights.  The challenges and potential opportunities for promising are presented individually as follow. 
\subsection{Challenges}
\textbf{Generalizability and transferability.} Significant progress has been achieved in developing predictive models with impressive performance. However, ensuring their generalization to unseen data and transferability across diverse chemical domains remains a formidable challenge. \emph{Multitask learning (MTL)} offers a promising solution by training models on multiple related tasts simultaneously. Yet, further exploration is needed to fully captilize on its potential to enhance generalization and transferability across different chemical domains.\\ 
\textbf{Data quality and representation.} High-quality, comprehensive molecular datasets are essential for training reliable predictive models. However, challenges include limited access to experimental data, constraints related to intellectual property and privacy. Standard property datasets contains only a limited number of molecules, which pose challenges for training advanced models. The success of generalizing to new chemical spaces is also heavily influenced by the quality of the training dataset. When the new data deviates considerably from the previous training set, it becomes more challenging for the model to accurately predict the target attribute. Therefore, a balance between dataset size and quality is essential for robust and effective model performance in new and challenging scenarios.\\ 
\textbf{Interpretability.} Achieving high predictive accuracy is essential, but understanding the factors driving model predictions is equally critical. While GNNs are being effectively utilize molecular graph representations, their decision-making process remain opaque. GNNs demonstrated superior performances over traditional ML methods in predicting various several molecular properties, including toxicity~\cite{li2021mutagenpred,shan2022predicting} and Lipophilicity~\cite{tian2023predicting}. Despite these successes, comprehending the rationale behind model predictions remains challenging, necessitating further investigation to bridge this gap in DL applications. Specifically, two key areas require attention: 1) Uncertainty estimation, which measures the degree of prediction reliability; and 2) Transparency, which involves knowing the process by which a system reaches a particular conclusion. Therefore, an interpretable model helps experts pinpoint performance issues and gain insights for future development.\\
\textbf{Multimodal Integration.} Molecular data is intrinsically heterogeneous and consists of multiple dimensions including chemical descriptors, molecular structures, and biological data. Combining these dimensions extracts the diverse molecular structure information from chemical compound~\cite{guo2020graseq}, which helps address sparse data issues in individual modality by compensating with data from other modalities. Novel approaches are needed to effectively integrate different types of molecular data—chemical, biological, and structural—with a particular emphasis on enhancing strategies for 3D structure analysis~\cite{karim2019toxicity, chen2022general, tang2022merged}.
By incorporating diverse data types, multimodal integration enhances the performance of predictive models by providing a more comprehensive understanding of molecular behavior.  
However, this integration also introduces challenges, particularly in dealing with the increased complexity and dimensionality of the data. Directly concatenating diverse modalities like text and image data can lead to representation conflicts and training imbalances, where sequential and discrete text data contrasts with the spatial and continuous nature of images. This can hinder model performance as it may bias towards one modality over another, potentially under utilizing information from each modality.\\
\subsection{Opportunities}
\textbf{Multitask Learning and Uncertainty Quantification in Molecular Modeling.} In multitask learning, training models across multiple tasks enhances predictive robustness by filtering out noise and biases specific to individual datasets. Beyond accuracy, understanding prediction certainty is crucial for researchers in molecular reasoning and experimental design, helping assess prediction trustworthiness. Despite extensive studies on uncertainty estimation techniques~\cite{busk2021calibrated,yang2023explainable}, consensus on the optimal approach for quantifying uncertainty in machine learning remains elusive. Effective uncertainty quantification varies with task and dataset specifics, requiring tailored methods. Establishing benchmark datasets that mimic diverse real-world scenarios is crucial for facilitating accurate comparisons and advancing uncertainty quantification methods.\\
\textbf{Advanced Learning Approaches.}
In exploring advanced learning paradigm, meta-learning~\cite {hospedales2021meta} approaches enable models to efficiently learn from a small number of molecules and generalize to new chemical spaces or properties. Few-shot learning complements this by integrating data augmentation techniques to generate additional training examples from limited data. Synthetic data generation methods, including molecular structure generation algorithms or property prediction simulations, further bolster dataset diversity and size. Additionally,  federated learning as highlighted by studies \cite{rieke2020future} and \cite{zhu2022federated} facilitates collaborative model training across decentralized data sources while preserving data privacy and security.\\
\textbf{Transparent appproaches in MPP.}
The benefits of an interpretable model are significant, aiding stakeholders in pinpointing root issues and suggesting future development directions when models perform poorly. ExplainableAI (XAI) techniques aim to offer explanations for model predictions. These techniques include methods such as feature importance scores, saliency maps, and attention mechanisms to clarify prediction mechanism and identify key factors.  While studies~\cite{henderson2021improving,gao2021gnes} have been explored explaining GNN outputs, substantial advancement in property prediction~\cite{sun2023explainable} are still needed.\\ 
\textbf{Efficiency and Representation Enhancement in MPP.} Contrastive learning can potentially enhance the efficiency and representation learning by learning to distinguish between similar and dissimilar pairs of data, thereby improving feature representations and ensuring the model effectively leverages information from all modalities~\cite{he2023novel}. Attention mechanisms also offer promising direction for dimensionality reduction through focus on the most relevant parts of the cross modal input data. Attention mechanisms can dynamically focus on the most relevant parts of the input data, effectively reducing the dimensionality by filtering out less important features. In contrast to early fusion, where different types of data are combined and fed directly into the final classifier~\cite{cai2022fp, wang2023investigating}, this approach uses fusion bottlenecks to limit the exchange of information between modalities at the latent level. This forces the model to distill and focus on the most relevant inputs from each modality. By ensuring that only essential information is shared it can lead to performance gains with reduced computational requirements~\cite{nagrani2021attention}.

Addressing these challenges requires interdisciplinary collaborations between researchers in computational chemistry, ML, statistics, and data visualization. Advanced techniques for feature selection, dimesionality reduction, regularization, and model interpretation are crucial for developing robust and efficient predictive models. Fusion techniques such as normalization, balanced training strategies, and modality-specific encoders are essential to effectively integrate and leverage the strengths of multiple modalities. 

%% file: Conclusion.tex
\section{Conclusion}\label{sec:conclusion}
This comprehensive survey aims to guide researchers through the current landscape of MPP, offering a foundation for future advancements. We discussed the different representations for molecules and provide an overview of encoding schemes, detailing the preprocessing steps necessary for transforming raw data, such as SMILES and molecular structures, into model inputs. We present a taxonomy for modality-based MPP, categorizing methods based on single and multiple modalities. We also explore the construction and training decisions of standard DL models and the prevalent learning schemes used to enhance MPP performance. Additionally, we identify popular benchmark datasets and tools for feature generation, presenting the top methods reported in the literature for each dataset to provide insights into model efficacy. Finally, we address the significant challenges in the field and outline future directions, highlighting both opportunities and areas needing further exploration. 

%% file: main.bbl

%% file: appendix.tex
\appendix
\section{Appendix}
\subsection{Neural Network Architectures}

\subsubsection{Graph Neural Networks}
GNNs represent a powerful class of neural network architectures tailored to process graph-structured data. Typically, GNNs involve three primary functions: \textbf{Message Passing:} Nodes in the graph share information with their neighbors iteratively via message passing. This operation entails aggregating information from surrounding nodes by using a weighted sum or a learned aggregation function. \textbf{Node Update:} After gathering information from neighbors, each node updates its own representation using the pooled data. This update step involve adding a neural network layer or a nonlinear activation function to the aggregated data. \textbf{Readout or Graph Level Operations:} To compute a graph-level embedding or predictions at the graph level, representations of all nodes are aggregated to form a final feature vector. We detail about the various variants of GNNs in the subsequent section. The illustration about the convolutional, attentional and message-passing in GNNs are depicted in Figure~\ref{fig:three GNNs}.

\begin{enumerate}
   
\item \textbf{GCN:} Among GNNs, GCNs\cite{zhang2019graph} stand out as a subtype that propagates information via node connections and takes into account both local and global structural information. They achieve this by generating embeddings for each node through the aggregation of information from its neighboring nodes. This aggregation process is crucial for capturing local structural features effectively. Specifically, the information obtained from neighboring nodes is weighted based on the inverse square root of the product of the degrees of the connected nodes. The message-passing operation in GCNs is subdivided into convolution operations and neighbor aggregation.\\
\begin{itemize}
        \item \textbf{Convolution operation:} The convolution operation in GCNs aggregates information from neighboring nodes. Given the input node features $h_v^l$ at layer $l$ for node $v$ and the adjacency matrix $A$  representing the graph structure, the messages for a node $v$ can be computed  using Equation \ref{gcn_mp}, where 
$deg(u)$, $deg(v)$ are the degrees of nodes $u$ and $v$, respectively.  
\begin{equation}
\label{gcn_mp}
    m_{v}^{l} = \sum\limits_{u\in neighbor(v)}^{} \frac{1}{\sqrt{deg(u)}\sqrt{deg(v)}}\cdot h_{v}^{l}
\end{equation}

   \item \textbf{Neighbor aggregation:} After computing the aggregated messages $m_{v}^{l}$, a weight matrix $W$ is used to perform linear transformation and aggregation as shown below with $\sigma$ as activation function such as ReLU.
   \begin{equation}
   \label{gcn_agg}
    z_{v}^{(l)} = \sigma\left(m_v^l\cdot W^{l}\cdot h_{v}^{l}\right)
    \end{equation}
\end{itemize}
Finally, the updated node features $h_v^{l+1}$ are computed by combining the aggregated neighbor information with the node's own features as in Equation \ref{gcn_upd} with $W_0$ as another weight matrix.
\begin{equation}
   \label{gcn_upd}
    h_{v}^{(l+1)} = \sigma\left(\sum z_v^l + W_0 \cdot h_v^l\right)
    \end{equation}
By performing multiple graph convolutional layers, GCNs capture hierarchical representations of molecular structures which are then used to predict various molecular properties such as solubility, toxicity, or bioactivity\cite{xiong2019pushing}. GCNs are trained using labeled data by adjusting their parameters to minimize the difference between predicted and actual property values, thus providing a powerful framework for accurate property prediction in computational chemistry\cite{ding2022combining, maron2019provably, wang2019graph}.\\
\item \textbf{GAT:} GATs\cite{velivckovic2017graph} are another type of GNN architecture used in MPP. GCNs has limited ability to capture fine-grained structural relationships within graphs. This is because it rely on fixed-weight aggregation functions to combine information from neighboring nodes, which may not adequately capture the varying importance of different nodes in influencing the central node's representation. Therefore, GATs introduce attention operation to address this challenge. 
\begin{itemize}
    \item \textbf{Attention operation :} It dynamically compute attention coefficients $\phi_{uv}$ to weight the contributions of neighboring nodes based on their relevance to the central node\cite{dinesh2023benchmarking}. The updated node features $h_v^{l+1}$ are computed via a weight matrix $W$, neighbors of node $v$ and attention score $\alpha_{uv}$ as shown in Equation \ref{gat}. The mathematical functions for $\alpha_{uv}$ and $ \phi_{uv}$ is given in Equation \ref{attention} and \ref{phi}, respectively. To ensure that the computed attention scores are comparable and stable across different nodes in the graph, attention scores are normalized using Equation \ref{attention}.
\begin{equation}
      \label{gat}
    h_{v}^{l+1} = \sigma \left(\sum\limits_{u\in neighbor(v)} \alpha_{uv}^{l} W \cdot h_{u}^{l}\right)
\end{equation}
 In MPP, GATs have shown promising results by effectively using attention mechanisms to learn informative node representations\cite{cai2022fp}. GATs offer greater flexibility in modeling complex graph structures which makes them well-suited for tasks where capturing fine-grained interactions between nodes is crucial for accurate predictions\cite{ahmad2023attention}.

\begin{equation}
     \label{attention}
    \alpha_{uv}^{l} = \frac {exp\left(\phi_{uv}^{l}\right)}{\sum\limits_{k\in neighbor(v)}exp \left(\phi_{vk}^{l}\right)}
\end{equation}

\begin{equation}
 \label{phi}
    \phi_{uv}^{l} = Attention\left(h_{v}^{l}\cdot h_{u}^{l}\right)
\end{equation}
 \end{itemize}
\begin{figure}[ht!] 
\centering
\includegraphics[scale = 0.75]
{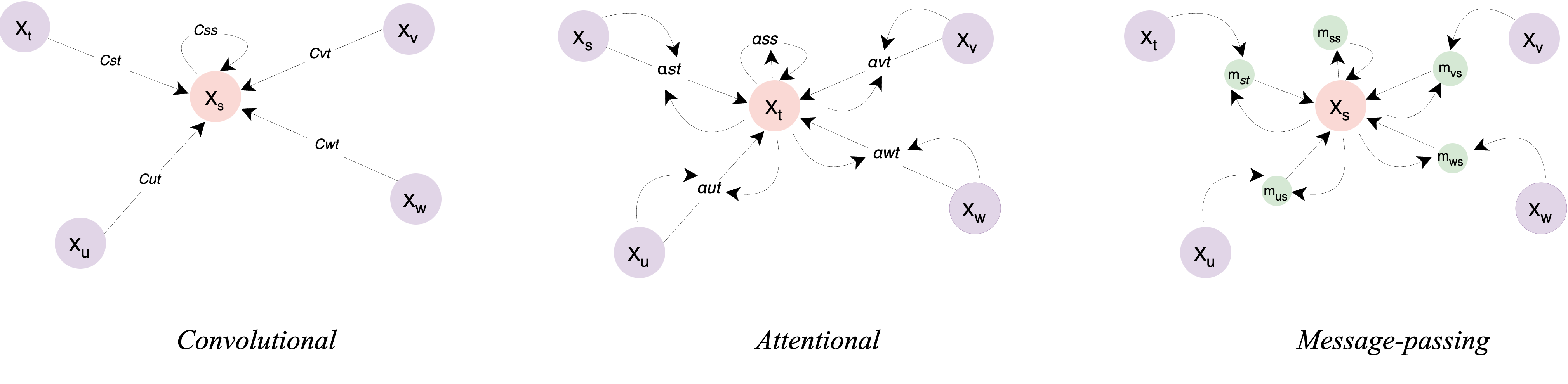}
\caption{Illustration showcasing the operations in GNNs. (a) Convolutional: allows nodes to communicate and exchange information with their neighbors in the graph. (b) Attentional:  determines the importance or relevance of neighboring nodes' features during message passing. (c) Message-Passing: encompasses the entire process of aggregating information from neighboring nodes and updating node representations based on this aggregated information.}
\label{fig:three GNNs}
\end{figure}
\item \textbf{MPNN:} Unlike GCN and GAT, MPNNs generalize the aggregation process by allowing custom message functions $M$ and update rules $U$, leading to potentially better performance on tasks that require capturing intricate node and edge relationships. It works iteratively with repeated message passing and updating steps. The computation of message and the update function is done as shown in Equation \ref{message} and \ref{update_mpnn}, respectively. One of the key advantages of MPNN in MPP is the flexibility to tailor the message passing and update mechanisms to specific tasks or domains. For example, different tasks may require different types of node interactions, and MPNNs allow these to be encoded directly into the model architecture\cite{tian2023predicting}. Additionally, MPNNs can be augmented with attention mechanisms or other forms of relational reasoning to selectively focus on important regions of the molecular graph\cite{busk2021calibrated}.
\begin{equation}
\label{message}
    M_{i}^{(l+1)}=\sum\limits_{j\in neighbor(i)}^{} M^{(l)}\left(h_{i}^{(l)}, h_{j}^{(l)}\right)
\end{equation}

\begin{equation}
\label{update_mpnn}
    h_{i}^{(l+1)} = U^{(l)}\left(h_{i}^{(l)}, M_{i}^{(l+1)}\right)
\end{equation}

\item \textbf{GIN:} To address the challenge of identifying graphs with the same structure but different node orderings which is often encountered in MPP tasks, GIN is used. Unlike traditional GNNs, GIN\cite{xu2018powerful} adopts a unique approach by bypassing explicit message passing mechanisms and graph convolutions. Instead, it directly embeds the neighborhood of each node using a learned aggregation function. This distinctive strategy empowers GIN to capture intricate structural patterns and relationships within molecular graphs more effectively, thereby enhancing its performance. 

\begin{equation}
      \label{GIN}
    h_{i}^{l+1} = MLP\left(\left( 1 + \epsilon^{l}\right)\cdot h_{i}^{l} + \sum\limits_{j\in neighbor(i)} h_{j}^{l}\right) 
\end{equation}
Here, $h_{i}^{l}$ represents the node embedding for node v at layer k, and $\epsilon^{l}$ is a learnable parameter for layer l. GIN collects information from neighbors by adding their embeddings and incorporating it into the updated node embedding using a MLP. The $\epsilon^{l}$ term adds a modest correction to avoid over-smoothing. Moreover, GIN exhibits inherent permutation invariance, and generates consistent embeddings regardless of the node ordering within the graph. This attribute renders GIN resilient to variations in graph topology which is particularly advantageous in MPP where molecules exhibit diverse sizes and structures\cite{kosasih2021graph, ahmad2023attention}. 

\begin{enumerate}
    \item The choice of message passing mechanism depends on the characteristics of the molecular data and the specific task requirements. For example, attentional mechanisms are effective for capturing long-range dependencies, while convolutional operations are suitable for capturing local structural patterns. The three different message passing mechanisms - Convolutional, Isomorphism, and Attentional can be explored based on the requirement. 
    \item After the message passing phase, a readout operation can be done using different methods. Some common readout operations include: 
    \begin{itemize}
    \item \textbf{Sum:} The node embeddings are simply summed to obtain the graph representation.
    \begin{equation}
        h_{graph} = \sum \limits_{v\in V}^{} h_v
    \end{equation}
    \item \textbf{Mean:} Similar to the sum readout, but instead of summing, the mean of all node embeddings is computed. 
    \begin{equation}
        h_{graph} = \frac{1}{|v|} \sum \limits_{v\in V}^{} h_v
    \end{equation}
    \item \textbf{Max:} The maximum value of each dimension across all node embeddings is taken to obtain the graph representation.
    \begin{equation}
        h_{graph} = \max \limits_{v\in V}^{} h_v
    \end{equation}
    \item \textbf{Attention-based readout:} An attention mechanism can be applied to assign importance weights to node embeddings before aggregating them.
     \begin{equation}
        \phi_{v} = \sigma \left( MLP \left(h_v\right)\right)
    \end{equation}
    \end{itemize}
    The selection of the readout operation influences the final graph-level representation and consequently the performance of the model in downstream tasks. 
    \item The architecture of GNN may includes multiple layers of message passing units. The decision regarding the number of layers, activation functions, hidden layer dimensions, and optimization methods significantly impacts the model's capacity and learning capability. Experimentation and tuning to determine the optimal architecture for the given MPP task is also essential.
\end{enumerate}
\end{enumerate}
\subsubsection{Recurrent Neural Networks}
RNNs are well-suited for modeling sequential data which makes them suitable for tasks involving molecules. RNNs have been applied in MPP tasks to analyze molecular sequences such as SMILES represented as linear sequences of characters or tokens. By processing input sequences incrementally and maintaining hidden states that retain information from previous steps, RNNs can effectively encode the structural and chemical characteristics of molecules, facilitating accurate property prediction. 
\begin{enumerate}
    \item \textbf{LSTM:} Unlike traditional feedforward neural networks, RNNs are designed to handle sequential data by incorporating hidden states that preserve information from earlier steps. However, RNNs often encounter the vanishing gradient problem which hampers their ability to capture long-term dependencies effectively. To mitigate this issue, LSTM networks were introduced to improve performance in capturing and retaining long-range dependencies in sequential molecular data\cite{su2019architecture, li2021smiles, hou2022accurate}. This is accomplished through the set of layers known as cell states $c_t (i_t, f_t, o_t)$ (Equation \ref{cell state}) and hidden state $h_t$ (Equation \ref{hidden state})which collectively enable the network to better understand the intricate structural relationships within molecules and retain essential information throughout the sequence.
\begin{equation}
\label{cell state}
    c_t = f_t \cdot c_{t-1} + i_t \cdot c_{t}^{'} 
\end{equation}

\begin{equation}
\label{hidden state}
    h_t = o_{t} \cdot tanh\left(c_{t}\right)
\end{equation}

where $c_{t}^{'}$ is the updated cell state given by Equation  \ref{updated cell state}. 
\begin{equation}
\label{updated cell state}
    c_{t}^{'} = tanh\left(x_{i}\cdot w_{ic} + w_{hc} \cdot h_{t-1} + b_{ic}\right)
\end{equation}
The decision regarding the handling of input, output, and retention of the previous cell state at each time step \emph{t} in the each cell is governed by the input gate $i_{t}$, output gate $o_{t}$ and forget gate $f_{t}$, respectively. The input, output, and forget states are computed using Equations \ref{input state}, \ref{output state}, and \ref{forget state}, respectively.
\begin{equation}
\label{input state}
    i_{t}^{} = tanh\left(x_{i}\cdot w_{ii} + w_{hi} \cdot h_{t-1} + b_{ii}\right)
\end{equation}

\begin{equation}
\label{output state}
    o_{t}^{} = tanh\left(x_{i}\cdot w_{io} + w_{ho} \cdot h_{t-1} + b_{io}\right)
\end{equation}

\begin{equation}
\label{forget state}
    f_{t}^{} = tanh\left(x_{i}\cdot w_{if} + w_{hf} \cdot h_{t-1} + b_{if}\right)
\end{equation}
\item \textbf{GRU:} The GRU represents a variant of recurrent networks akin to the LSTM model. While maintaining similar functionality to LSTM, GRU exhibits a somewhat simpler architecture. GRUs are more computationally efficient compared LSTM networks which makes them suitable for large-scale MPP tasks involving extensive datasets of molecular structures\cite{hu2020deep}. While there are fewer studies employing GRUs in MPP compared to other architectures, their simpler design and reduced parameter count render them more amenable to training and deployment, particularly in resource-limited settings. One notable distinction is that GRU consolidates both the hidden state and cell state into a unified hidden state, thereby reducing the parameter count. Similar to LSTM, GRU incorporates gating mechanisms to regulate information flow within the network. It is characterized by two gating components: the reset gate $r_t$ and the update gate $z_t$ as given below in Equation \ref{gru reset state} and Equation \ref{gru update state}. These gates enable GRU to selectively retain and update information over time and contribute to its effectiveness in modeling sequential data, including property prediction tasks.
\begin{equation}
\label{gru reset state}
    r_{t}^{} = Activation\left(x_{i}\cdot w_{ir} + w_{hr} \cdot h_{t-1} + b_{ir}\right)
\end{equation}
\begin{equation}
\label{gru update state}
    z_{t}^{} = Activation\left(x_{i}\cdot w_{iz} + w_{hz} \cdot h_{t-1} + b_{iz}\right)
\end{equation}
The reset gate in GRU regulates the amount of past information to discard, while the update gate manages the integration of new information. GRU introduces a "candidate hidden state" $h_{t}^{'}$ computed using the reset gate to selectively update the hidden state based on the importance of past and new information (Equation \ref{gru hiddden state}). This mechanism enhances the model's capability to retain relevant historical context while adapting to new inputs, leading to improved performance in sequential tasks like language modeling and time series prediction.

\begin{equation}
\label{gru hiddden state}
    h_{t}^{'} = Activation\left(x_{it}\cdot w_{ih} + w_{hr} \cdot \left(r_t * h_{t-1}\right) + b_{ih}\right)
\end{equation}
The final hidden state $h_{t}$ is computed as a linear interpolation between the previous hidden state  $h_{t-1}$ and the candidate hidden state $h_{t}^{'}$, with the weighting controlled by the update gate (Equation \ref{final hidden gru}). This interpolation mechanism allows the model to determine the extent to which the new candidate state should replace the previous state at each time step, facilitating the retention of relevant information while updating the hidden representation.
\begin{equation}
\label{final hidden gru}
    h_{t} = \left( 1-z_{t} * h_{t-1}\right) + z_t * h_{t}^{'}
\end{equation}\\
\end{enumerate}

\subsubsection{Transformers}
The transformer model, originally designed for sequence-to-sequence tasks like machine translation and text summarization in NLP, has emerged as powerful models in cheminformatics tasks. 
The transformer model introduced by Vaswani et al.~\ref{vaswani2017attention}revolutionized sequence processing by leveraging attention mechanisms, paving the way for parallel computation of input sequences. Unlike RNNs, which are inherently sequential, transformers enabled simultaneous processing, leading to notable reductions in computation time and the ability to handle longer sequences, capturing more extensive dependencies. The attention mechanism is fundamental to this architecture, relying on three linear transformations known as query (Q), key (K), and value (V) vectors, each with a length of $d_k$. These vectors undergo dot product computations, facilitating attention calculations for every input sequence. For example, given a sequence of input vectors $X = \left( x_1, x_2, \dots, x_n \right)$, the self-attention mechanism computes new representations $Z = \left( z_1, z_2, \dots, z_n \right)$. The attention mechanism is as shown in Equation \ref{transformer1}. Transformers use a self-attention mechanism to analyze input data in parallel, making them particularly useful for sequential and structured data\cite{maziarka2020molecule}. They use multiple attention heads\cite{song2023double}, which allows the model to focus on different portions of the input data at the same time as shown in Equation \ref{transformer2}. This improves the model's ability to capture diverse patterns and relationships between different parts of a molecule, enabling more accurate property prediction.
 \begin{equation}
 \label{transformer1}
     Attention\left( Q, K,V \right) = softmax \left( \frac{QK^{T}}{\sqrt{d_k}}\right)\cdot V
 \end{equation}
 \begin{equation}
 \label{transformer2}
     MultiHead\left( Q, K, V \right) = Concat\left( head_1, head_2, \dots, head_n\right)\cdot W_O
 \end{equation}

 Here Q, K, and V are the input matrices, each representing query, key, and value, $d_k$ is the dimension of the key vector, $head_i = Attention\left( QW_{Qi}, KW_{Ki}, VW_{Vi}\right)$ is the $i^{th}$ attention head and $W_O$ is the output linear transformation. The transformer model integrates layer normalization for stable computations and a feed-forward neural network to introduce non-linearity, alongside the attention mechanism. Tokenized sequences are processed by the encoder unit, which focuses on token relationships using self-attention, while the decoder unit generates predictions based solely on previous tokens. This architectural design not only reduces computational costs and captures longer dependencies but also forms the basis for pre-training and fine-tuning strategies. Overall, transformers represent a promising direction for MPP, offering state-of-the-art performance, scalability, and flexibility in modeling diverse molecular structures and properties. For transformer based studies please refer section 3. As research in this area continues to advance, further developments in transformer-based approaches are expected to drive significant progress in MPP.